\documentclass{article}

\usepackage{arxiv}

\usepackage[utf8]{inputenc} %
\usepackage[T1]{fontenc}    %
\usepackage{hyperref}       %
\usepackage{url}            %
\usepackage{booktabs}       %
\usepackage{amsfonts}       %
\usepackage{nicefrac}       %
\usepackage{microtype}      %
\usepackage{lipsum}		%
\usepackage{graphicx}
\usepackage{natbib}
\usepackage{doi}
\usepackage{enumitem}
\usepackage{algorithm}
\usepackage{algpseudocode}
\usepackage{amsmath}
\usepackage{wrapfig}
\usepackage{minted}
\usepackage[most]{tcolorbox}

\usepackage{tikz}
\usetikzlibrary{arrows.meta, positioning}

\usepackage[textsize=tiny]{todonotes}
\usepackage{caption}
\usepackage{listings}
\usepackage{xcolor}               %
\usepackage{amsmath}
\usepackage{pgfplots}
\pgfplotsset{compat=newest}
\usepackage{subcaption}
\usepackage[capitalize,noabbrev]{cleveref}

\title{
\includegraphics[height=1.8cm]{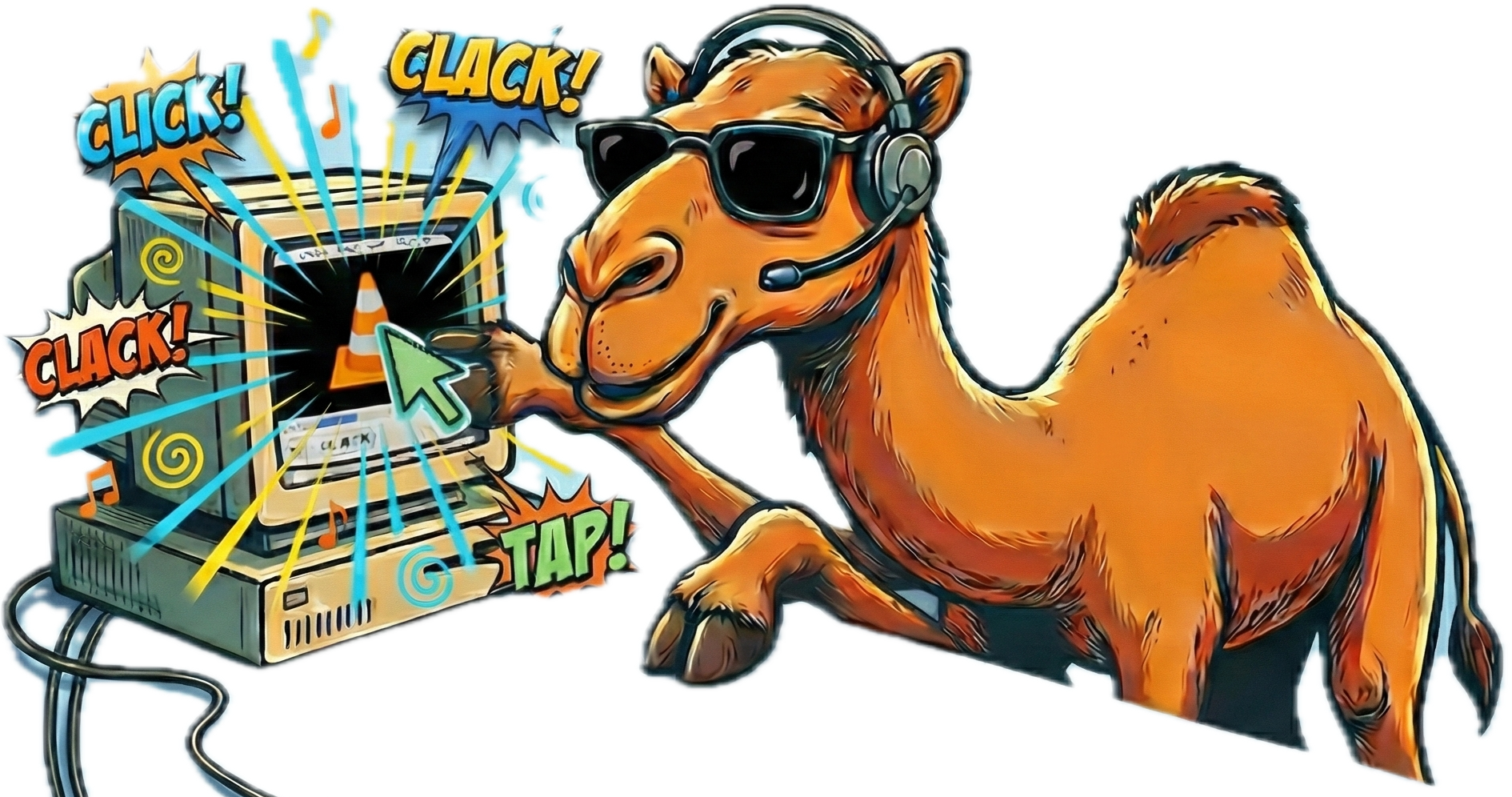}
CaMeLs Can Use Computers Too:\\System-level Security for Computer Use Agents}

\author{
    Hanna Foerster$^{*1}$ \quad
    Tom Blanchard$^{*2}$ \quad
    Kristina Nikolić$^{3}$ \quad
    Ilia Shumailov$^{4}$ \quad
    Cheng Zhang$^{4}$ \\[0.2em]
    \textbf{Robert Mullins$^{1}$} \quad
    \textbf{Nicolas Papernot$^{2}$} \quad
    \textbf{Florian Tram\`{e}r$^{3}$} \quad
    \textbf{Yiren Zhao$^{4}$} \\[0.5em]
    \small
    $^{*}$Equal contribution \\
    \small
    $^{1}$University of Cambridge \quad
    $^{2}$University of Toronto \& Vector Institute \quad
    $^{3}$ETH Zurich \quad
    $^{4}$AI Sequrity Company
}

\date{}

\renewcommand{\shorttitle}{\includegraphics[height=0.5cm]{content/figures/camel_cua_logo.png} System-level Security for Computer Use Agents}

\hypersetup{
pdftitle={A template for the arxiv style},
pdfsubject={q-bio.NC, q-bio.QM},
pdfauthor={David S.~Hippocampus, Elias D.~Striatum},
pdfkeywords={First keyword, Second keyword, More},
}

\begin{document}
\maketitle

\begin{abstract}
AI agents are vulnerable to prompt injection attacks, where malicious content hijacks agent behavior. Among proposed defenses, architectural isolation provides the strongest guarantees by strictly separating trusted task planning from untrusted environment observations. However, applying this design to Computer Use Agents (CUAs), which automate tasks by viewing screens and executing actions, presents a fundamental challenge. Current agents require continuous observation of UI state to determine each action, which conflicts with the isolation required for security. We resolve this tension by demonstrating that UI workflows, while dynamic, are structurally predictable. Single-shot planning, where a trusted planner emits upfront a complete branching plan covering all anticipated runtime states, provides control flow integrity guarantees against arbitrary instruction injections. We introduce \textbf{NOVA} (\textit{Navigating via Observation, Verification, and Action}) to make this viable in the combinatorially large UI state space, where the plan can invoke a perception model to resolve runtime values such as UI coordinates. We evaluate our design on OSWorld, and retain up to 57\% of the performance of frontier models while improving performance for smaller open-source models by up to 19\%, demonstrating that rigorous security and utility can coexist in CUAs. Although upfront planning prevents instruction injections, we show that additional measures are needed to defend against \textbf{Branch Steering} attacks, where adversaries deceive the perception model into routing execution down attacker-preferred branches of the plan, such as redirecting the agent to a malicious website.\footnote{Code is available at: \href{https://github.com/cleverhans-lab/camel-cua}{https://github.com/cleverhans-lab/camel-cua}} 
\end{abstract}

\section{Introduction}

Computer Use Agents (CUAs) are large Vision-Language Model (VLM)-based systems that automate tasks by perceiving user interfaces through screenshots or structured data and executing actions.
Unlike text-based agents with well-defined tool APIs, CUAs use both semantically ambiguous tools and loosely constrained action parameters. While text-based tools like \texttt{send\_email(to, subject, body)} have clear semantics, CUA actions like \texttt{click(x, y)} are context-dependent: the same action could log in, delete data, or navigate to a malicious site depending on the target coordinates.
This ambiguity creates attack surfaces where malicious instructions can be embedded directly into UI content via prompt injections and adversarial pixel perturbations, with real-world exploits demonstrating consequences including data exfiltration and arbitrary code execution~\citep{Li2025Commercial}.
Consequently, ensuring control flow integrity, which guarantees that the agent only executes actions prescribed by a user, has become a paramount security requirement.

To secure agentic systems, recent literature proposed to adopt a system-centric approach: the Dual-LLM paradigm~\citep{willison2023dual-llm,Costa2025,Debenedetti2025}. This architecture separates planning (a trusted Privileged Planner, or P-LLM, blind to the environment) from perception (a Quarantined Perception model, or Q-VLM, that processes untrusted observations within the planner's plan), structurally enforcing control flow integrity. CaMeL~\citep{Debenedetti2025} is one such system, where the P-LLM emits a complete execution plan upfront before any environment observation. 
While effective in simulated agentic benchmarks such as AgentDojo~\citep{debenedetti2024agentdojo}, where application state is represented as structured Python objects and agents interact through a closed set of typed tool APIs (e.g., \texttt{send\_email}, \texttt{get\_calendar}), applying this pattern to CUAs presents a fundamental challenge. The state space a CUA planner must anticipate grows combinatorially across application state, UI layouts, and pixel-level variations, rendering it effectively unbounded unlike the finite, enumerable state spaces of prior Dual-LLM evaluations in text-only environments with typed tool APIs. Standard CUAs manage this complexity through a data-dependent, multi-turn loop, perceiving the UI state at every step to determine the next action. Strict isolation blocks malicious injections by preventing the planner from observing the screen, but simultaneously appears to sever the visual feedback loop these systems depend on.

\begin{figure*}[t]
    \centering
    \begin{subfigure}[t]{0.05\textwidth}
        \vspace{0pt}
        \centering
        \includegraphics[width=\textwidth]{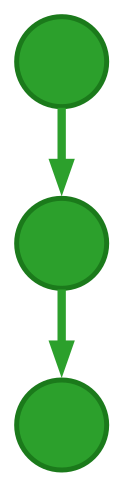}
        \caption{CaMeL-AgentDojo}
        \label{fig:exec_agentdojo}
    \end{subfigure}
    \hfill
    \begin{subfigure}[t]{0.06\textwidth}
        \vspace{0pt}
        \centering
        \includegraphics[width=\textwidth]{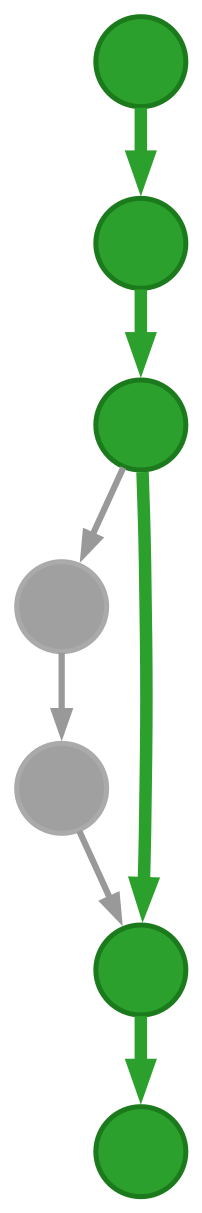}
        \caption{CaMeL-CUA-unoptimized}
        \label{fig:exec_unoptimized}
    \end{subfigure}
    \hfill
    \begin{subfigure}[t]{0.25\textwidth}
        \vspace{0pt}
        \centering
        \includegraphics[width=\textwidth]{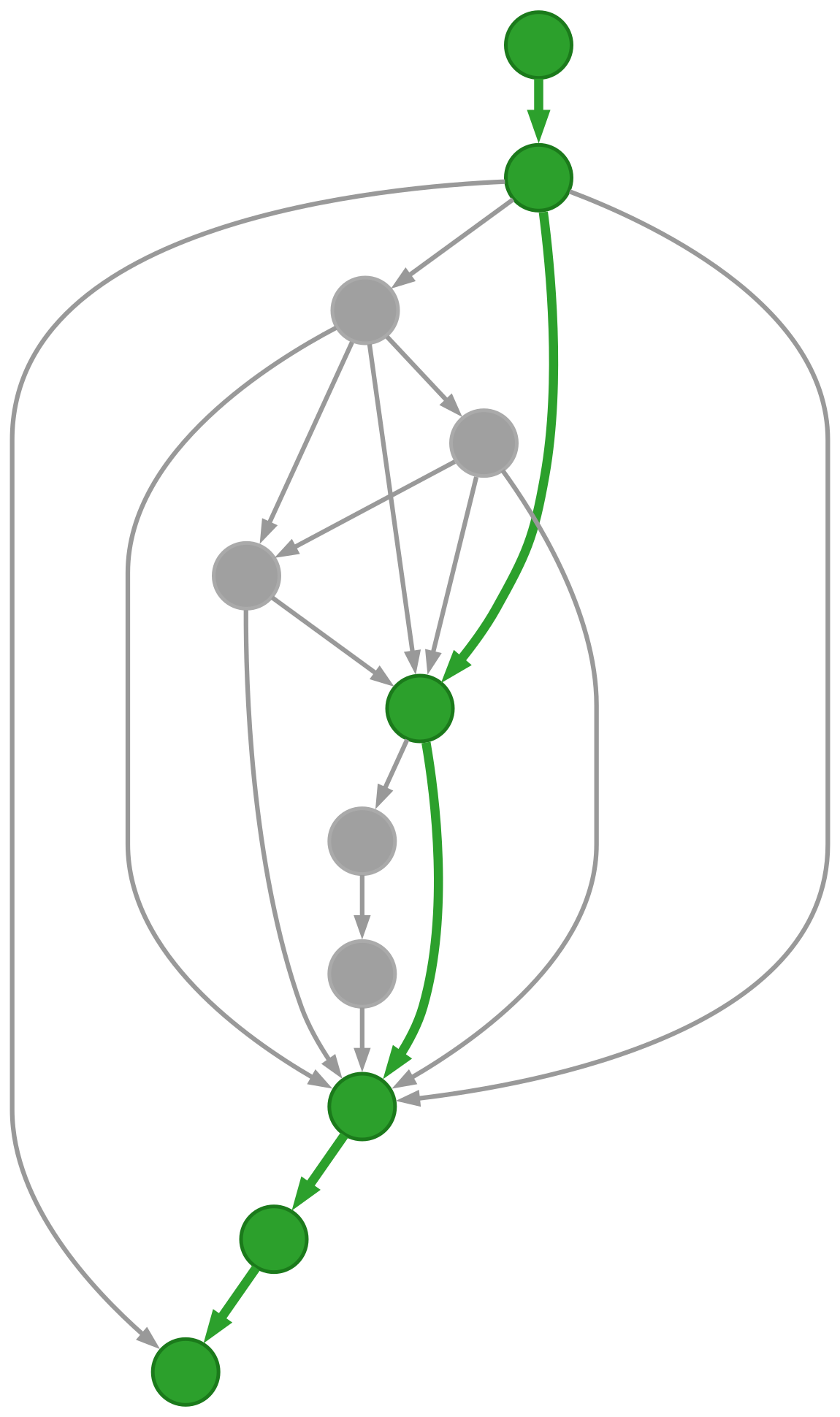}
        \caption{CaMeL-CUA-NOVA}
        \label{fig:exec_optimized}
    \end{subfigure}

    \vspace{0.8em}
    \footnotesize
    \begin{tabular*}{\textwidth}{@{\extracolsep{\fill}} l r r r}
    \toprule
                  & \textbf{CaMeL-AgentDojo} & \textbf{CaMeL-CUA-unoptimized} & \textbf{CaMeL-CUA-NOVA} \\
    \midrule
Length        & $51.8 \pm 6.9$  & $71.6 \pm 3.7$  & $213.3 \pm 7.5$ \\
Tool Calls    & $4.9 \pm 0.4$   & $19.8 \pm 1.7$  & $41.1 \pm 1.6$  \\
Branches      & $3.7 \pm 0.5$   & $11.3 \pm 0.8$  & $39.7 \pm 1.7$  \\
Jaccard Sim.  & 0.393           & 0.001           & 0.044           \\
    \bottomrule
    \end{tabular*}

    \caption{Execution path analysis with aggregate plan statistics (dataset-wide means), viewed in a graph structure. Green nodes show a single execution trace; grey nodes are untaken contingency branches. \textit{Length} = mean code lines, \textit{Tool Calls} = mean function calls, \textit{Branches} = mean conditional branches, \textit{Jaccard Sim.} = intra-task pattern similarity (lower is more diverse). We compute values with their 95\% CI (confidence interval). CUA tasks demand substantially more complex plans than AgentDojo (more tool calls, more branches, and far lower intra-task similarity), and CaMeL-CUA-NOVA produces significantly richer contingency exploration than the unoptimized baseline. More statistics about CaMeL plans for the 3 evaluated setups can be found in \cref{tab:plan_stats,tab:plan_stats_per_suite}, and detailed plans in \cref{fig:graph_comparison}.}
    \label{fig:exec_paths}
\end{figure*}

To our knowledge, we present the first demonstration that Dual-LLM isolation can be successfully adapted to CUAs. Our key insight is that at any given execution step, the runtime states an agent could be in are predictable enough for a planner to enumerate the corresponding fallbacks upfront, even when the underlying state space is unbounded. Prior Dual-LLM settings such as AgentDojo had enumerable state and typed tool outputs, so plans were largely linear with shallow rather than deeply nested branching. CUA plans cannot rely on this structure and must fan out into nested branches covering combinatorially many runtime states the planner cannot observe. This forces an order of magnitude more tool calls and branches than AgentDojo plans, with roughly half of plan nodes serving as untaken fallback paths (\cref{fig:exec_paths}). We therefore introduce NOVA (\textit{Navigating via Observation, Verification, and Action}), a planning methodology built around a \texttt{verify\_hypothesis} primitive, and call the CaMeL-based instantiation \textbf{CaMeL-NOVA}. Each branch proceeds in two quarantined steps before acting. First, a Q-VLM call gathers an observation of the current screen. Then \texttt{verify\_hypothesis}, a Q-LLM call under a fixed comparison prompt, semantically checks that observation against a predicted condition (e.g., ``is a cookie banner visible?'') and returns a boolean verdict. The interpreter uses this verdict to select a pre-written branch, executing the encoded action (e.g., a click at coordinates returned earlier by the Q-VLM). Control flow is thus determined entirely by the plan: adversaries cannot inject instructions that deviate from it. CaMeL-NOVA adds the structural contingency planning that makes CaMeL viable for CUAs (\cref{fig:exec_paths}). Our OSWorld evaluation shows that this CFI-guaranteed CUA still retains significant utility relative to standard multi-turn CUAs, with performance improving substantially under stronger planner models and multiple independent planning attempts.

\begin{figure*}[t]
    \centering
    \begin{subfigure}[b]{0.36\textwidth}
        \centering
        \includegraphics[width=\textwidth]{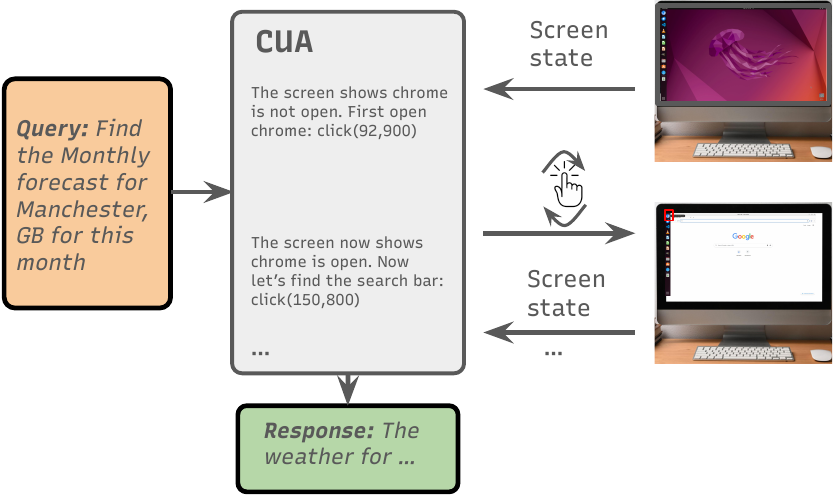}
        \caption{Standard CUA with iter. observation}
        \label{fig:standard_cua}
    \end{subfigure}
    \hfill
    \begin{subfigure}[b]{0.6\textwidth}
        \centering
        \includegraphics[width=\textwidth]{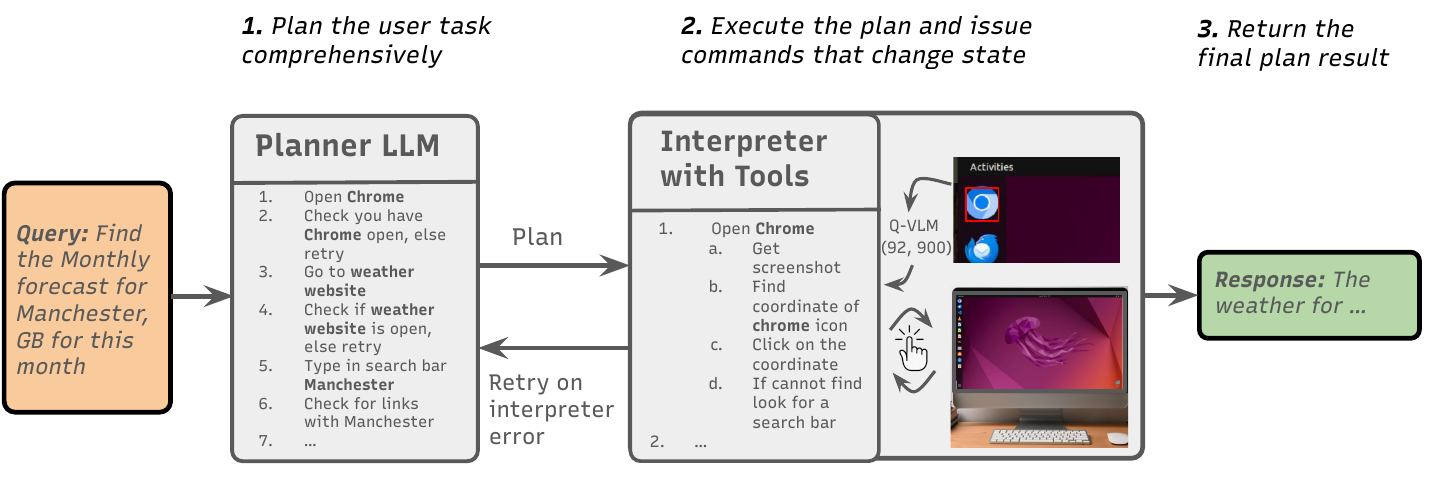}
        \caption{Our Dual-LLM CUA with single-shot planning}
        \label{fig:camel_cua_visualization}
    \end{subfigure}
    \caption{Contrasting standard CUA with our Dual-LLM adaptation. Left: every turn exposes the planner to untrusted observations, enabling instruction injection at any step. Right: the P-LLM emits a complete branching plan before any observation. The interpreter executes the plan, invoking the Q-VLM to resolve environment-dependent values (e.g., UI coordinates from \texttt{find}) that are bound to plan variables and consumed by subsequent actions. Attackers are restricted to pre-written paths.}
    \label{fig:architecture_comparison}
\end{figure*}

Finally, while our architecture guarantees control flow integrity, we show it retains a critical vulnerability inherent to data-dependent systems: Branch Steering. Traditional prompt injection attacks exploit CUAs by injecting new actions that were never part of the user's intended task, such as emailing credentials to an attacker during a benign browsing session. 
Our architecture prevents such arbitrary command injection, but Branch Steering exploits a different surface: attackers manipulate visual cues (e.g., fake buttons or concealed Document Object Model or DOM elements) to deceive the perception model. The perception model then returns false information that routes execution down an attacker-preferred, yet structurally valid, path in the pre-written plan. For instance, the agent might click a malicious advertisement disguised as a legitimate cookie popup. 
We therefore implement redundancy-based verification mechanisms: cross-checking screenshot and DOM consistency, and seeking consensus from independent models. We evaluate their effectiveness against both prompt-injection and pixel-based branch steering attacks.

In summary, this work makes the following contributions:
\begin{itemize}
    \item \textbf{Plan-Complexity Gap Between Typed-API Agents and CUAs:} We measure how plans differ between AgentDojo and CUA settings: CUA plans need an order of magnitude more tool calls and branches than AgentDojo plans, with roughly half of plan nodes serving as fallback paths that do not execute on any given run (\Cref{fig:exec_paths,tab:plan_stats}). The gap motivates the architecture in our second contribution.
    
    \item \textbf{Dual-LLM Architecture for CUAs (CaMeL-NOVA):} The first Dual-LLM architecture for Computer Use Agents, using Single-Shot Planning with the NOVA paradigm and a \texttt{verify\_hypothesis} primitive that lets plans branch on runtime observations without exposing them to the planner, providing Control Flow Integrity guarantees. CaMeL-NOVA achieves up to 19\% performance improvement for small open-source models and retains up to 57\% of performance for larger closed-source models on the OSWorld benchmark.
    
    \item \textbf{Branch Steering and Defenses:} We identify \textit{Branch Steering} as a distinct data-flow threat vector, where attackers manipulate visual cues (e.g., fake buttons) to fool the agent into choosing a dangerous, yet valid, path within its pre-written plan. We demonstrate its feasibility, evaluate redundancy-based mitigation, and highlight the fundamental distinction between control-flow and data-flow security in isolated architectures.
\end{itemize}

\section{Background}
\label{sec:background}

\subsection{Computer Use Agents}
\label{sec:background:CUA}
\textbf{CUA Capabilities and Architectures.}  Computer Use Agents (CUAs) require \textit{perception} (understanding UI state through VLMs processing screenshots or structured data like DOM trees), \textit{action} (executing from a generalizable action space), \textit{reasoning} (task decomposition and planning), and \textit{memory} (execution history and task knowledge) \citep{qin2025ui}. Architectures range from end-to-end VLMs \citep{qin2025ui,wang2025opencuaopenfoundationscomputeruse,fu2025mano,ye2025mobile} to agentic frameworks with specialized modules \citep{Agent-S,Agent-S2,Agent-S3,song2025coact1computerusingagentscoding} (see \cref{sec:app:more_about_cuas}). Both paradigms rely on iterative, reactive planning, continuously adapting plans based on real-time environmental feedback, in contrast to the upfront single-shot planning we adopt in this work. 

\textbf{CUA Attacks. }Previous attacks on CUAs achieve arbitrary control over task execution by injecting malicious instructions or optimized pixels into text and visual content. Attack vectors include adversarial pop-ups \citep{zhang2025attacking}, malicious image patches \citep{aichberger2025mip}, fine-print injections \citep{chen2025obvious}, and redirection attacks via trusted platforms like Reddit \citep{Li2025Commercial}, with benchmarks like OS-HARM \citep{kuntz2025osharm} and WASP \citep{evtimov2025wasp} demonstrating goals such as data exfiltration and arbitrary code execution. 
These attacks assume a reactive planner exposed to untrusted content; our threat model instead asks which attacks survive when the planner is architecturally isolated from the environment.

\textbf{CUA Defenses. }
Existing defenses for CUAs rely on pattern recognition rather than architectural isolation and are known to be fragile \citep{nasr2025attackermovessecondstronger}. \citet{yang2025incontext} use few-shot learning on malicious and benign examples to recognize attacks, but fail on novel ones. \citet{hu2025agentsentinel}'s AgentSentinel monitors actions via predefined rules and an LLM auditor, but the auditor itself is vulnerable to poisoning. 
CeLLMate enforces HTTP-level sandboxing policies, a natural fit for browser agents but difficult to adapt to CUAs \citep{meng2025cellmate}. CUA action spaces are broader (spanning OS dialogs and desktop applications) and operate at the visual level (clicks at pixel coordinates), with no equivalent protocol layer to intercept. None of these defenses provide architectural isolation of the planner from untrusted content, the property our Dual-LLM CUA adaptation targets.

\subsection{Dual-LLM}

\textbf{Dual-LLM and System-Centric Security.} Two instantiations of Dual-LLM have been proposed for text-based agents. CaMeL transforms user queries into formal plans with explicit access constraints enforced by an interpreter \citep{Debenedetti2025}. Fides instead calls the planner iteratively, generating one action per turn while redacting tool-call outputs from the P-LLM \citep{Costa2025}. 
While both approaches provide strong guarantees about \textit{which} actions can be executed, they face fundamental limitations with data-dependent workflows where the required action sequence itself is specified in untrusted data. (See \cref{sec:extendedbackground} for more details.) In this paper, we adapt both to CUAs.

\textbf{What Exactly Does Dual-LLM Mitigate?} 
We distinguish two security properties. \textit{\textbf{Control flow integrity (CFI)}} ensures the structure of execution -- which function calls run, in what order, and under what conditional logic -- is fixed by the planner. \textit{\textbf{Data flow security}} ensures untrusted data cannot manipulate the arguments to those calls or the values driving conditional decisions. Dual-LLM guarantees CFI: branching logic and action sequences are determined entirely by the planner, which never observes the environment. It does not prevent data-flow attacks, where malicious content shapes the Q-LLM's outputs and thereby the arguments or conditional values that flow through the plan. Prior work addresses data-flow attacks via semantic policies and capabilities, but these require tools with intrinsic meaning; CUA actions like \texttt{click} have none without environmental context, making policy-based protection difficult to deploy.

\textbf{Branch Steering Attacks.} 
Prior evaluations of prompt-injection defenses focus on control-flow attacks that execute actions outside the user's intent \citep{shi2025lessonsdefendinggeminiindirect,nasr2025attackermovessecondstronger}. These attacks are structurally impossible under Dual-LLM, since planning and execution are separated. We therefore introduce \textit{Branch Steering}, a distinct attack class that targets data flows within a P-LLM plan: an adversary manipulates environmental observations (e.g., via adversarial images or DOM injection) to trigger a valid but malicious conditional branch in the pre-approved plan. In our setup, such attacks act through environment-querying functions such as \texttt{verify\_hypothesis} and \texttt{find}.

\section{Methodology}

\textbf{Threat model.} We assume an attacker who cannot access or modify the P-LLM but can inject malicious content into portions of the rendered environment (e.g., advertisements, forum posts, attacker-controlled websites). The attacker knows the defender's task and toolset and can approximately predict the P-LLM's plans, a worst-case assumption under which defenses effective against this attacker remain effective under weaker ones. For tractability we restrict to browser-based tasks. Full threat model and justification of attacker-knowledge assumptions is in \cref{sec:app:threat_model}.

\subsection{CaMeL-NOVA: Adapting Dual-LLM to Computer Use}
\label{sec:method:camel_nova}

We adopt the Dual-LLM architecture, separating responsibilities into two distinct components. We formalize this section based on CaMeL, while Fides is a minor expansion where the plan is decoded one instruction at a time. Both CaMeL and Fides originally included security policy components to restrict dataflow; we exclude such policies, so our setup represents a worst-case Dual-LLM security baseline since we do not perform additional use-case-specific policing. The two components are:

\begin{itemize}
    \item \textbf{Privileged Planner (P-LLM)}: A secure LLM that generates a complete execution plan single-shot before any interaction occurs. It has no access to the live environment.
    
    \item \textbf{Quarantined Perception (Q-VLM)}: An untrusted Vision-Language Model that interacts with the environment (screenshots/DOM). It executes within the P-LLM's plan and cannot modify its structure.
\end{itemize}

Because the P-LLM cannot see the screen to adjust its plan dynamically, we bridge the semantic gap using a comprehensive toolset allowing for conditional execution. The P-LLM compiles the user query into a Python-like plan utilizing specific functions that for example summarize the screenshot state, return coordinates, or check the state of the environment with a screenshot or the DOM.
A high-level overview of how Dual-LLM works is visualized in~\cref{fig:camel_cua_visualization}, and the core of our secure execution strategy is formalized in Algorithm~\ref{alg:camel_cua}, along with example plans in \cref{sec:app:example_plans}. 

To enable effective single-shot plans without intermediate reasoning, we enforce an \textbf{Observe-Verify-Act} methodology in the P-LLM system prompt. The P-LLM writes calls to Q-VLM functions at plan time but never observes their return values; the interpreter invokes these calls at execution time and binds the outputs to plan variables consumed by subsequent actions and branches.

\begin{itemize}
    \item \textbf{Observe}: The plan issues Q-VLM calls that aggregate state information (visual summaries or DOM elements).
    \item \textbf{Verify}: The plan calls \texttt{verify\_hypothesis}, which uses a Q-LLM under a fixed comparison prompt to check a predicted condition against the current Q-VLM observation and return a boolean (e.g., ``\textit{Am I on the login page?}''). The interpreter uses this boolean to select the next branch.
    \item \textbf{Act}: The plan executes interactions (clicks/types) only after verification.
\end{itemize}

This structure allows the P-LLM to reason in advance by generating branches and loops centered around  \texttt{verify\_hypothesis} outcomes, anticipating multiple failure modes and state transitions. We arrived at this methodology by analyzing successful plans on OSWorld: they consistently gathered state information first, explicitly verified assumptions, and only then acted. This pattern proved especially critical for handling the ambiguous initial states in benchmark tasks, where one-sentence descriptions provide no information about the current computer environment. 

Observe-Verify-Act is an instance of a broader design principle: not every aspect of a CUA plan needs to be rediscovered per task. Routines such as handling cookie banners, detecting whether a browser is already open, or recovering from a failed click appear in virtually every browser-based task, regardless of the user's goal. We treat these as \emph{shared plan components} and encode them directly in the P-LLM system prompt: how to structure Observe-Verify-Act branches, how to handle mandatory UI interactions, when to fall back to alternate find strategies. This frees the P-LLM to reason only about the \emph{task-specific components}, such as which website to visit, which element carries the relevant information, how to combine steps into a workflow. This separation identifies what CUA planning needs to solve dynamically versus what can be compiled into the agent framework once and reused. Such engineering discoveries are orthogonal to our security guarantees (a more developed system prompt improves utility without weakening Control Flow Integrity), but necessary to make an architecture this restrictive practically useful. Closing the utility gap to demonstrate feasibility required engineering effort on par with prior agentic CUA frameworks \citep{Agent-S,song2025coact1computerusingagentscoding}. Further implementation details are in \cref{sec:extended-methods}.

\subsection{Characterizing the Residual Data-Flow Attack Surface}
\label{sec:residual_attack_surface}
Control flow integrity does not prevent data-flow manipulation: a compromised Q-VLM can return false coordinates or state summaries that steer the plan into attacker-chosen branches, an attack class we call \textit{Branch Steering}. Since semantic policies do not transfer to CUA tools (\cref{sec:background}), any data-flow defense for CUAs will necessarily be probabilistic. We anticipate the two verification strategies most likely to emerge in practice and stress-test them to characterize the residual attack surface: \textbf{DOM Consistency}, which cross-checks Q-VLM outputs against the ground-truth DOM, and \textbf{Multi-Modal Consensus}, which uses an independent VLM to verify the Q-VLM's output against both the screenshot and the original instruction. Both leverage the empirically low transferability of adversarial examples between diverse models \citep{schaeffer2024failurestransferableimagejailbreaks}. Details are in \cref{sec:app:redundancy_details}.

We develop two Branch Steering attacks, each characterizing a distinct weakness class in probabilistic data-flow defenses (illustrations in \cref{fig:cookie_attack,fig:pixel_attack}; full details in \cref{sec:app:further_details_attack}). \textbf{1) Cookie Popup Prompt Injection} characterizes the limits of DOM-based verification by showing that attackers can rewrite what the DOM communicates: we embed fake cookie popups inside ad banners that redirect the agent to attacker-controlled sites when \texttt{find} is called to locate cookie popups. Because cookie handling is a predictable early step that the planner treats as a generic repeatable routine, each \texttt{find(``cookie popup'')} call resolves independently of the agent's location in the plan, enabling multi-step variants through benign hop sites that bypass hyperlink verification and a long-range variant placing the fabricated popup on a later webpage the adversary predicts the agent will visit. \textbf{2) Pixel Attack} characterizes the limits of Multi-Modal Consensus by showing that the Q-VLM output itself can be adversarially induced to agree with verifier expectations: gradient-based pixel perturbations on ad banners make the Q-VLM return plausible thought traces and coordinates aligned with the element search instruction rather than the banner's own coordinates (which would be flagged), so screenshot, DOM, and Q-VLM output all appear legitimate.

\section{Evaluation}
\label{sec:eval}

We evaluate three questions:
(1) what is the structural gap between CUA plans and typed-API agent plans,
and how does NOVA's planning methodology reshape this structure
(\cref{tab:plan_stats,fig:exec_paths});
(2) how well does Dual-LLM retain utility on complex computer-use tasks
(\cref{tab:vlm_security_comparison,tab:planner_model_comparison}); and
(3) how do our redundancy defenses hold up against Branch Steering attacks.

We conduct our evaluation on OSWorld~\citep{xie2024osworld}, a benchmark for measuring agent performance on realistic computer use tasks across applications including Chrome, LibreOffice, and GIMP, with automated verification scripts. Following standard practice in CUA benchmarking, we report pass@k metrics, which measure the probability that at least one correct solution appears in k independent attempts at a task. Pass@k represents an upper bound assuming oracle selection, characterizing the performance achievable if we could pick the best of k candidate plans. Our architecture supports developing such selectors since plan generation is unconditional on prior failures and can therefore run in parallel. We also report pass@1 as a conservative estimate of single-attempt performance.

\textbf{Experimental Setup.} For our experiments, we integrate the OSWorld benchmark~\citep{xie2024osworld} into official code-bases of CaMeL\footnote{\url{https://github.com/google-research/camel-prompt-injection}} and Fides\footnote{\url{https://github.com/microsoft/fides}}. We set up three different end-to-end CUAs, of which two are small open-source models---UITars-1.5-7B~\citep{ui-tars-15-seed}, OpenCUA-32B~\citep{wang2025opencuaopenfoundationscomputeruse}, and one is a large closed source model---Claude Sonnet 4.5~\citep{anthropic2025claudesonnet45}. Further, we test 9 different frontier LLMs as our planner models, of which three are open-source---Deepseek-R1-0528~\citep{guo2025deepseek}, Kimi-K2-Thinking~\citep{team2025kimi}, GPT-OSS-120b~\citep{agarwal2025gpt}, while 6 are closed source---Claude Sonnet 4.5~\citep{anthropic2025claudesonnet45}, Gemini 2.5 Pro~\citep{comanici2025gemini}, Gemini 3 Pro Preview~\citep{google2025gemini3}, GPT-5~\citep{singh2025openai}, GPT-5.1~\citep{singh2025openai}, Grok-4~\citep{xai2025grok4}.

\textbf{Baseline utility.} Without Dual-LLM isolation, our results align with published OSWorld benchmark numbers with UITars-1.5-7B, and OpenCUA-32B achieving 24.4\%, and 28.7\% respectively at a 15-step cap (per-task breakdown in \cref{sec:app:base_utility} \cref{tab:task_distribution}). Our Claude results (37.7\%) are substantially lower than the published performance of 58.1\%, which we attribute to our significantly reduced step count (15 vs. 50 steps), enforced due to time and cost limitations.

\textbf{CUA tasks are substantially harder than AgentDojo tasks.} AgentDojo's bounded, enumerable state space lets plans stay short: 4.9 tool calls and 51.8 code lines per plan. Most branches (79.4\%) are deterministic control flow over plan-internal values, since the planner can enumerate the possible cases at plan time without runtime verification. For CUA tasks, neither property holds: the state space is unbounded across application state, UI layouts, and pixel-level variations, and runtime UI behavior cannot be predicted at plan time. Even the unoptimized CaMeL-CUA baseline (UITars$^\dagger$) needs 19.8 tool calls and 71.6 code lines per plan to navigate this; NOVA's Observe-Verify-Act scaffolding pushes this to 41.1 tool calls and 213.3 code lines. Branch counts grow even more sharply (3.7 / 11.3 / 39.7) because each anticipated UI state typically needs its own fallback path; data-flow edges follow the same pattern (\cref{tab:plan_stats}). Plan similarity across different samplings of plans for the same task tells a parallel story, where shared scaffolding helps. AgentDojo plans overlap substantially (Jaccard 0.393) because the tasks are simple and common tool patterns repeat across them. The unoptimized baseline drops to 0.001, with no shared scaffolding to leverage. NOVA brings sharing back to 0.044 by encoding the repeating routines (cookie handling, browser-state checks, fallback finds) in the system prompt, so the same scaffolding shows up across tasks and the planner focuses on task-specific structure that can lead to pass@k recovery. More details about the metrics used for this comparison can be found in \cref{app:sec:plan_analysis_methodology}.

\textbf{NOVA's branching structure provides fault tolerance, which closes the utility gap.} NOVA plans average close to one branch per tool call (39.7 per 41.1), whereas the unoptimized baseline has only 11.3 branches and 19.8 tool calls per plan (\cref{tab:plan_stats}; example plan in \cref{sec:app:example_plans:subsec:natural_products}). Without \texttt{verify\_hypothesis} or system-prompt instructions about which situations need verification, the planner either skips verification and fallbacks entirely or substitutes brittle structural checks on Q-VLM returns. Errors then cascade rather than triggering a recovery branch. Empirically, the unoptimized baseline only succeeds on short, mostly self-contained sequences (e.g., changing a font size, installing a browser extension), and fails on tasks requiring multiple sub-goals or many dependent edits. UITars with CaMeL-NOVA closes a 40\% gap at pass@3 (\cref{tab:vlm_security_comparison}).%

\textbf{Planner quality dominates utility.} On UITars's 60-task set, UITars, OpenCUA, and Claude reach 65.0\%, 66.7\%, and 68.3\% pass@5 under CaMeL-NOVA. The three are within 3.3 points of each other despite very different CUA backends. Performance is therefore primarily driven by the planner. \cref{tab:planner_model_comparison} compares 9 frontier planners on a 17-task subset: GPT-5 leads (12/17) followed by Grok-4 (10/17). Weaker planners fail in two ways: their plans omit fallback branches for unexpected UI states, and they lack robust multi-step strategies for navigating complex applications. Gemini 3 Pro, Gemini 2.5 Pro, and Claude Sonnet 4.5 additionally produce highly similar plans across runs at default sampling. Additional samples then rarely recover from earlier failures, so pass@k flattens (example plans in \cref{sec:app:example_plans:subsec:natural_products,sec:app:example_plans:subsec:natural_products_gemini3}).

\textbf{Fides reaches comparable utility at much higher cost.} Fides-NOVA achieves 33.3\% pass@1 and 66.7\% pass@5 on the 60-task UITars set (\cref{sec:app:fides_utility,tab:fides_pass3_results}), confirming that NOVA generalizes across Dual-LLM substrates. The overhead profile, however, differs sharply. CaMeL-NOVA adds 1.88$\times$ tokens over the undefended agent, while Fides-NOVA adds 29.6$\times$ because it re-invokes the planner with the full history at every execution step rather than once. Adding Multi-Modal Consensus on top of CaMeL-NOVA raises the total to 6.57$\times$ (\cref{app:token_count}).

\textbf{Both attacks succeed against our redundancy defenses.} \textit{1) Cookie Popup Attack:} the one-step attack succeeds with no defense. DOM Consistency blocks it on static Google banners via the ``Advertisement'' frame in the DOM, but fails on HTML5 banners, which replace the advertisement frame with a cookie frame. Multi-Modal Consensus struggles to distinguish fabricated cookie banners from legitimate ones without degrading utility, consistent with weak success rates of prompt-based defenses \citep{evtimov2025wasp}. \textit{2) Pixel Attack:} even under our strongest redundancy defense (Multi-Modal Consensus), configured to heavily prioritize security over utility with high false positive rates, the pixel attack evades both screenshot-based and DOM-enhanced Q-VLM verification (\cref{fig:pixel_attack}). Such attacks could be exploited by advertisers (e.g., pharmaceutical companies) to redirect CUA traffic to their product pages. We leave multi-objective and multi-Q-VLM pixel attacks to future work (details in \cref{app:cookie_attack_details,app:pixel_attack_details}).

\textbf{These failures reflect a fundamental coverage-utility tradeoff in probabilistic data-flow defenses.} Using Claude-Haiku-4.5 as the DOM checker and GPT-5 as the screenshot checker for Multi-Modal Consensus, both prompted to detect advertisement-banner injections, we measure flagging rates on cookie-mimicry attacks (10 runs) and false-positive rates on benign tasks (the 17 planner-evaluation tasks; \cref{sec:app:redundancy_eval,tab:redundancy_utility_evaluation}). Narrow targeting bounds false positives but misses broader attack vectors, and broadening coverage incurs significant utility loss, consistent with \citet{zhang2025attacking}.

\begin{table*}[t]
\centering
\caption{Performance comparison of CUA backends with \textbf{CaMeL-NOVA} across different task sets. NOVA (\emph{Navigating via Observation, Verification, and Action}) is a structured planning method we introduce, in which the planner is prompted to organize each step as Observe$\rightarrow$Verify$\rightarrow$Act. We first contrast the CaMeL-CUA-unoptimized baseline (a generic system prompt without Observe$\rightarrow$Verify$\rightarrow$Act guidance or other CUA navigation guidance, and no \texttt{verify\_hypothesis}; denoted UITars\textsuperscript{$\dagger$}), against UITars with CaMeL-NOVA on the same 60 tasks: 11/60 (18.3\%) vs 35/60 (58.3\%) at Pass@3 --- a 40-point gap demonstrating that NOVA's planning method is essential for the CaMeL+CUA setup to work. We then evaluate CaMeL-NOVA across different CUA backends (OpenCUA-32B and Claude Sonnet 4.5) on the same task set and find that CUA differences become less significant under CaMeL-NOVA. OpenCUA on OpenCUA tasks and Claude on Claude tasks show similar results even at larger set sizes, underscoring the strong performance of CaMeL-NOVA. For All tasks, we evaluate only with UITars as the CUA at a lower pass@k rate and include the 30 non-automatically evaluable tasks that are originally set to success by default on OSWorld in the Overall row (second value, $n=369$), and find that CaMeL-NOVA performance (29\%) yields a higher success rate than just UITars (24.4\%).}
\label{tab:vlm_security_comparison}
\resizebox{\textwidth}{!}{
\begin{tabular}{lrrrrrrr}
\toprule
\textbf{Category} & \multicolumn{4}{c}{\textbf{UITars Tasks}} & \textbf{OpenCUA Tasks} & \textbf{Claude Tasks} & \textbf{All Tasks} \\
\cmidrule(lr){2-5} \cmidrule(lr){6-6} \cmidrule(lr){7-7} \cmidrule(lr){8-8}
 & \textbf{UITars\textsuperscript{$\dagger$}} & \textbf{UITars} & \textbf{OpenCUA} & \textbf{Claude} & \textbf{OpenCUA} & \textbf{Claude} & \textbf{UITars}\\
\midrule
Chrome              & 1/11 (9.1\%)   & 6/11 (54.5\%)  & 8/11 (72.7\%)  & 8/11 (72.7\%)  & 11/13 (84.6\%) & 8/14 (57.1\%)  & 10/46 (21.7\%)\\
GIMP                & 2/6 (33.3\%)   & 5/6 (83.3\%)   & 4/6 (66.7\%)   & 4/6 (66.7\%)   & 5/9 (55.6\%)   & 6/11 (54.5\%)  & 5/16 (31.3\%)\\
LibreOffice Calc    & 0/3 (0.0\%)    & 2/3 (66.7\%)   & 2/3 (66.7\%)   & 2/3 (66.7\%)   & 2/3 (66.7\%)   & 4/13 (30.8\%)  & 6/46 (13.0\%)\\
LibreOffice Impress & 0/9 (0.0\%)    & 6/9 (66.7\%)   & 5/9 (55.6\%)   & 6/9 (66.7\%)   & 9/14 (64.3\%)  & 7/17 (41.2\%)  & 13/47 (27.7\%)\\
LibreOffice Writer  & 1/8 (12.5\%)   & 3/8 (37.5\%)   & 4/8 (50.0\%)   & 3/8 (37.5\%)   & 3/7 (42.9\%)   & 8/12 (66.7\%)  & 6/22 (27.3\%)\\
Multi-apps          & 1/5 (20.0\%)   & 3/5 (60.0\%)   & 3/5 (60.0\%)   & 3/5 (60.0\%)   & 3/3 (100.0\%)  & 4/7 (57.1\%)   & 11/100 (11.0\%)\\
OS                  & 0/4 (0.0\%)    & 3/4 (75.0\%)   & 3/4 (75.0\%)   & 4/4 (100.0\%)  & 4/7 (57.1\%)   & 7/11 (63.6\%)  & 8/19 (42.1\%)\\
Thunderbird         & 2/4 (50.0\%)   & 2/4 (50.0\%)   & 3/4 (75.0\%)   & 3/4 (75.0\%)   & 5/7 (71.4\%)   & 7/8 (87.5\%)   & 6/14 (42.9\%)\\
VLC                 & 0/2 (0.0\%)    & 1/2 (50.0\%)   & 1/2 (50.0\%)   & 1/2 (50.0\%)   & 1/4 (25.0\%)   & 0/4 (0.0\%)    & 2/17 (11.8\%)\\
VS Code             & 4/8 (50.0\%)   & 8/8 (100.0\%)  & 7/8 (87.5\%)   & 7/8 (87.5\%)   & 5/9 (55.6\%)   & 11/12 (91.7\%) & 10/18 (55.6\%)\\
\midrule
\textbf{Overall}    & \textbf{11/60 (18.3\%)} & \textbf{39/60 (65.0\%)} & \textbf{40/60 (66.7\%)} & \textbf{41/60 (68.3\%)} & \textbf{48/76 (63.2\%)} & \textbf{62/109 (56.9\%)} & \shortstack[r]{\textbf{77/339 (22.7\%),}\\\textbf{107/369 (29.0\%)}}\\
\midrule
\textbf{Pass@1}     & 4/60 (6.7\%) & 25/60 (41.7\%) & 22/60 (36.7\%) & 23/60 (38.3\%) & 22/76 (28.9\%) & 31/109 (28.4\%) & 51/339 (15.0\%)\\
\textbf{Pass@2}     & 8/60 (13.3\%)             & 30/60 (50.0\%) & 29/60 (48.3\%) & 30/60 (50.0\%) & 32/76 (42.1\%) & 46/109 (42.2\%) & 70/339 (20.6\%)\\
\textbf{Pass@3}     & 11/60 (18.3\%)             & 35/60 (58.3\%) & 35/60 (58.3\%) & 33/60 (55.0\%) & 37/76 (48.7\%) & 57/109 (52.3\%) & 77/339 (22.7\%)\\
\textbf{Pass@4}     & -              & 35/60 (58.3\%) & 37/60 (61.7\%) & 35/60 (58.3\%) & 44/76 (57.9\%) & 61/109 (56.0\%) & - \\
\textbf{Pass@5}     & -              & 39/60 (65.0\%) & 40/60 (66.7\%) & 41/60 (68.3\%) & 48/76 (63.2\%) & 62/109 (56.9\%) & - \\
\bottomrule
\end{tabular}
}
\end{table*}

\section{Discussion}
\label{sec:discussion}

Our investigation challenges the assumption that strict system-centric security is incompatible with GUI automation, and shows that CUA tasks are often less data-dependent than they appear. We establish a new baseline for secure CUAs, while exposing trade-offs between plan rigidity, data dependency, and residual attack vectors.

\textbf{What CFI Still Provides.} First, it eliminates the arbitrary-instruction injection class entirely: attacks that introduce actions the planner did not write are structurally impossible under Dual-LLM. Residual Branch Steering attacks are therefore constrained to manipulating actions the plan already contains -- when a plan relies on verification steps, the data returned by the Q-VLM effectively becomes a control signal for branch selection. As enumerated in \cref{app:cfi_attack_scenarios}, attack difficulty grows with both the number of coordinated Q-VLM manipulations required and the specificity of plan alignment needed, ranging from single-call manipulations against generic utility routines to coordinated multi-call compromise of task-specific plans like fund transfers. Second, an explicit plan enables a cluster of properties that continuous-observation agents struggle to provide: policy-level protections become tractable once actions and arguments are enumerated at plan time; plans can be inspected or approved before execution, enabling user oversight; and the plan is a stable interface that other defenses (e.g., browser-layer sandboxing \citep{meng2025cellmate}) can compose with. Of these, realizing policy-level protections requires translating coordinate-level actions into semantic ones, a natural direction for future work.

\textbf{Privacy.} The Dual-LLM split also enables privacy-preserving deployment: proprietary models handle planning without observing sensitive environment content, while open-source models run perception locally, reducing cost and preserving user privacy.

\textbf{Feasibility of Single-Shot Planning in (seemingly) Dynamic Environments.} Contrary to the expectation that strict control-data isolation would sever CUA functionality, our results demonstrate that a Dual-LLM architecture is viable. This finding suggests that current academic benchmarks rely less on real-time reactivity than previously thought: advanced general reasoning capabilities are sufficient to navigate complex app-specific tasks via careful planning. In our architecture, performance is primarily driven by the reasoning and memory capability of the P-LLM rather than the reactive capabilities of the Q-VLM. This suggests that as reasoning models improve, the utility of single-shot secure planning is likely to increase without requiring fundamental architectural changes.

\textbf{Task--Data Dependency.} While our \textit{Observe-Verify-Act} methodology allows for significant utility, we identify data dependency as the main limitation factor in the security-utility trade-off. The success of a single-shot plan depends on the P-LLM's ability to anticipate start states and failure modes; underspecified tasks (e.g., those lacking initial state descriptions) force the planner to account for exponentially more branches, degrading performance and increasing costs. However, we find that performance can be significantly improved through prompt tuning, suggesting that seemingly data-dependent tasks can often be reduced to data-independent ones by: (1) fine-tuning on the task distribution to increase predictability, (2) incorporating extensive state-verification logic (\texttt{verify\_hypothesis}) into the prompt to handle runtime ambiguity, and (3) explaining general reasoning behind planning.

\textbf{Better Benchmarks.} Many of the tasks found in OSWorld are ill-defined ("\textit{Increase font here"}), unmeasurable automatically \footnote{30 tasks in OSWorld, e.g., \url{https://github.com/xlang-ai/OSWorld/blob/cbc3b590ff7573034e5614ba74a4b1483a0e75b5/evaluation_examples/examples/chrome/ae78f875-5b98-4907-bbb5-9c737fc68c03.json}}, and are inherently data independent (e.g., "\textit{get to natural products at \url{drugs.com}}", where there is no explicit button to go there and the agent either needs to explore or know what precise buttons to click). 

\textbf{Reflection on Costs and Security Scaling.} We observe that pass@5 significantly improves performance. Because CaMeL samples plans unconditionally of prior failures, it is possible to sample multiple plans in parallel or combine them into a comprehensive super-plan. This indicates that CaMeL in CUAs scales favorably with model inference costs: as foundational models become more efficient, high-security control flow should become increasingly accessible. In fact, when running the pass@20 we observe that Claude scales even further in performance, as we show in~\Cref{fig:claude_scaling} in~\cref{sec:app:scaling_claude}, to approximately $73\%$ by pass@20.

\section{Conclusion}

Our work has successfully challenged a core assumption in AI security: that robust, system-centric defenses are incompatible with the dynamic, data-dependent nature of Computer Use Agents. By adapting the Dual-LLM framework with the \textit{Observe-Verify-Act} methodology, we have established a new and surprisingly effective baseline for secure CUA operation. Our results demonstrate that single-shot planning preserves significant utility, improving up to 19\% utility with smaller open-source models (pass@3) and retaining up to 57\% performance with larger closed-source models (pass@5) on CUA tasks, empirically validating the viability of secure CUA deployment. Importantly, this approach shifts the primary driver of success to the Privileged Planner's reasoning capabilities, meaning future utility improvements can leverage advances in foundational LLM capabilities without compromising the security architecture.

While Dual-LLM successfully isolates control flow, it remains vulnerable to our new \textit{branch steering attacks}, where a malicious environment coerces the Quarantined Perception model into returning data that forces execution into attacker-chosen paths within a legitimate plan. Even with additional defenses, residual vulnerabilities persist, as highlighted by the practical threat of our cookie attack.

\bibliographystyle{unsrtnat}
\bibliography{references}  

\newpage
\appendix
\section{Extended Background}
\label{sec:extendedbackground}
\subsection{Dual-LLM}
\label{sec:background:subsec:dual-llm}
Prior work shows that model-level defenses alone are insufficient against prompt injection attacks, motivating system-centric defenses that constrain the information a model is allowed to see to harden the \textbf{control flow}~\cite{wu2024system,willison2023dual-llm,wu2025isolategpt,zhong2025rtbas,abdelnabi2025firewalls}. We assume this setting and focus on Dual-LLM-style designs.

\begin{algorithm}[H]
\caption{Two Flavors of Dual-LLM}
\label{alg:dual_llm_comparison}
\footnotesize
\begin{algorithmic}[1]
\Statex \textbf{CaMeL (Static Planning):}
\State $\Pi \leftarrow \text{P-LLM}(\text{Task}, \text{Tools})$ \Comment{Generate full plan}
\For{step $s \in \Pi$}
    \State $\text{Interpreter}(s)$ \Comment{Enforce policy}
    \If{$s$ needs data}
        \State $\text{Q-LLM}(Data)$ \Comment{Isolated process}
    \EndIf
\EndFor
\Statex
\Statex \textbf{Fides (Iterative Planning):}
\While{Task not done}
    \State $H_{safe} \leftarrow \text{Redact}(History)$ \Comment{Hide values}
    \State $a_t \leftarrow \text{P-LLM}(H_{safe})$ \Comment{Next single step}
    \State $r_t \leftarrow \text{Exec}(a_t)$ \Comment{Save to var}
    \State $History \leftarrow History \cup (a_t, r_t)$
\EndWhile
\end{algorithmic}
\end{algorithm}

\textbf{Dual-LLM variants and their guarantees.} 
Dual-LLM separates control flow (P-LLM) from data flow (Q-LLM) and has been extended in two directions. CaMeL~\cite{Debenedetti2025} compiles a task into a typed, Python-like plan with capability metadata and an interpreter that enforces control flow and allows to add data-flow policies on tools. Fides~\cite{Costa2025} retains this dual-LLM separation but forgoes a static plan in favor of writing only a single line of the plan in each turn and saving outputs of tool calls, or Q-LLM calls to variables to use for later. Note that the content of these variables are redacted and cannot be seen from the viewpoint of the P-LLM. 
\textit{Under the condition that the Q-LLM is not given control over parts of the control flow}, these systems guarantee that environment information cannot derail task execution, as only function calls that were planned to be executed without environment knowledge can be executed. In particular, there are cases of tasks where the dual-LLM pattern cannot provide solutions: data-dependent tasks where the required sequence of actions is itself specified in untrusted data. For such tasks, the P-LLM cannot determine what steps to execute without seeing the untrusted content, which would violate the isolation guarantee. This represents a fundamental limitation of maintaining strict separation between planning and data processing.

For data independent tasks, outputs from the Q-LLM or tool-calls can be used for adding conditional logic or loops with conditionals into the plan and dealing with unknowns in the environment. However, even in this case, dual-LLM does not necessarily provide guarantees for \textbf{data flow} manipulations. An attacker could inject malicious content into the environment that influences the Q-LLM's outputs when it processes that untrusted data. Since the Q-LLM's extracted values are used as arguments to tool calls and branching logic, the attacker could achieve some damage with this, even while adhering to the P-LLM's plan. CaMeL addresses data flow security through \textit{security policies} and \textit{capabilities}. Capabilities are metadata tags tracking each value's provenance and allowed readers, enabling fine-grained restrictions on individual data items. Security policies are Python functions that check whether tool executions are permissible based on these capabilities, for instance, blocking emails sharing confidential documents with unauthorized recipients, even if those addresses were Q-LLM-extracted from prompt-injected sources. 

However, defining policies requires manual domain expertise with no automatic derivation method, which becomes a bottleneck for dynamic environments with a huge set of environment states. As most users are not experts in recognizing what policy allowances might incur which security risks, some policies need to be preset and others need to be automatically adaptive and personalized. The personalization is necessary, as users might have significant differences in preference of whether they want to share certain information or want the agent to take specific paths of execution in certain contexts~\cite{ghalebikesabi2024operationalizingcontextualintegrityprivacyconscious}. All of this taken together means that it cannot be a solely a manual human task to set policies but rather requires a model trained specifically for this. However, such policies may be incorrect, necessitating a worst-case baseline. Therefore, when attempting to adapt Dual-LLM to CUAs we do not attempt to add security policies but rather restrict how the Q-LLM can be called and add redundancy style defenses inside the Q-LLM tool call. We hope that this baseline can highlight the benefits of adding a Dual-LLM framework on CUAs, while also underlining remaining difficulties that we hope future work can build upon.

\subsection{Redundancy-based defenses}

Algorithm-Based Fault Tolerance (ABFT) is a concept from multiprocessor systems which adds redundancy and error detection into a system to enhance fault tolerance~\cite{huang1984_abft,banerjee1986concurrent,nair1990analysis}. Such redundancy trades off computation to add reliability, and upon error detection, selective recomputation or reconfiguration can be triggered. The core principle is N-version programming\cite{avizienis1985n,chen1978n}, where multiple independent implementations solve the same problem and their outputs are compared. Disagreements signal potential faults. In our setting, we treat function outputs that reveal information about the environment as potentially faulty nodes and employ redundancy through diverse modality inputs or changes in the prompt and/or different VLMs that process a query. This is analogous to data diversity techniques in fault-tolerant systems~\cite{ammann1988data_diversity}, where the same logical input is represented differently to expose errors that might be masked in a single representation. 

\subsection{How do CUAs work?}
\label{sec:app:more_about_cuas}

\textbf{Four capabilities needed for CUA performance: Perception, Action, Reasoning, Memory} \citep{qin2025ui}. \emph{Perception} refers to the component that provides real-time understanding of the environment by recognizing and interpreting the dynamic user interface. This understanding can be obtained from structured text provided by APIs such as HTML, the accessibility tree, or the Document Object Model (DOM), or from visual input such as screenshots, or from a combination of both. Structured text may present challenges due to verbosity and limitations depending on the operating system or application, while a VLM processing a screenshot requires the ability to identify elements in the environment, generate descriptions of these elements, read text using Optical Character Recognition (OCR), understand spatial relationships and element locations, capture state transitions, and support question answering. \emph{Action} refers to the capability to predict the next action within a defined action space. The action space should be designed by decomposing user tasks into subtasks and generalizing across components that are reused in multiple tasks. \emph{Reasoning} involves the processes required to decompose problems into subtasks, handle longer horizon dependencies, and support iterative refinement based on intermediate outcomes. Finally, the \emph{memory} module can be divided into short-term memory, which captures state transitions and maintains execution history, and long-term memory, which consists of prior knowledge as well as information acquired from previous tasks and interactions.

\textbf{Agent frameworks vs. end-to-end models. }CUA architectures can be broadly categorized into architectures using an agentic framework, which distributes tasks to specialized modules, and end-to-end agent models, which rely on a single model for all capabilities needed to gain utility in CUA tasks~\citep{qin2025ui}. 

\textit{Agentic frameworks} can have specialized VLMs for perception, action, reasoning and memory, and have seen success in benchmarks with CUAs such as Agent-S, and CoAct~\citep{Agent-S,Agent-S2,Agent-S3,song2025coact1computerusingagentscoding}. Notably, Agent-S2~\citep{Agent-S2} employs a compositional framework with a high-level manager which decomposes a task into a list of subgoals, a worker that generates actions to complete a subgoal in natural language, and grounding experts which receive this natural language instruction and turn them into UI actions. They employ proactive planning which reevaluates and updates the plan after every subtask. \citet{song2025coact1computerusingagentscoding} with CoAct-1 employ one central planner module called Orchestrator, which decomposes the user task and assigns subtasks to either a programmer agent or GUI operator agent. Basing off the insight that some actions can be achieved more reliably through programmatic measures, the programmer agent writes and executes python and bash scripts, whereas the GUI operator performs the conventional UI actions such as clicking. Finally, Agent-S3~\citep{Agent-S3} recognizes that paths towards solving a task can branch in unpredictable ways due to delayed feedback, mistaken reasoning, and the dynamic environment and can accumulate mistakes in a destructive manner. While the other two frameworks employed hierarchical planning, they only employ one end-to-end model but run three of them in parallel. Then they summarize the trajectories into narratives of how the task was solved and run a Best-of-N fourth run based on the best path. However, \citet{qin2025ui} and \citet{ye2025mobile} list adaptability to dynamic environments or unfamiliar tasks, dependence on manual optimization and expert knowledge, maintenance overhead, and disjoint learning paradigms between specialized modules as limitations.

Several works~\citep{qin2025ui,wang2025opencuaopenfoundationscomputeruse,fu2025mano,ye2025mobile} are examples of \textit{end-to-end CUA models}. They similarly formulate the problem of CUA task solving into a multi-step sequential planning problem, which consists of a \emph{system prompt}, a \emph{user task} given as the first \emph{user prompt}, and a multi-turn conversation between the VLM (\emph{assistant message}) and tool call outputs returned as a \emph{user message}. The content of this user message consists of the tool call, tool output, and screenshot after the tool was called. The assistant message always contains the tool calls and parameters to execute next, as well as depending on the model, context thoughts as to why this was chosen. Assistant message and user message interleave and create a conversation history until the assistant deems the task solved. Many CUAs are specifically trained VLMs that have been trained to excelling at perception, action, reasoning, and memory, with short-term memory often being tuned adhoc through conversation history length.

\section{Extended methodology section}
\label{sec:extended-methods}

\subsection{Combining Dual-LLM and CUA}
\label{sec:app:combining_dualllm_and_CUA}

\begin{algorithm}[htbp]
\footnotesize
\caption{CaMeL Adaptation for Computer Use Agents (CUA)}
\label{alg:camel_cua}
\begin{algorithmic}[1]
\Require User Task $T$, System Prompt $S_{prompt}$ (containing "Observe-Verify-Act" instructions), Environment $E$
\Require \textbf{P-LLM} (Privileged Planner), \textbf{Q-VLM} (Quarantined Perception), \textbf{Verifier} (Redundancy Check)
\Ensure Execution Status (Success/Fail)

\Statex \Comment{\textbf{Phase 1: Single-Shot Plan Generation} }
\State Define Toolset $\mathcal{F} \leftarrow$
\Statex \hspace{\algorithmicindent} $\{ \text{summarize\_screenshot}, \text{find}, \text{find\_element\_by\_text},$ 
\Statex \hspace{\algorithmicindent} $\text{verify\_hypothesis}, \text{check\_done} \}$
\State $S_{prompt} \leftarrow S_{prompt} \cup \text{Instructions}(\text{"Use Dual Finding Strategy"})$
\State $\Pi \leftarrow \text{P-LLM}(S_{prompt}, T, \mathcal{F})$ \Comment{Compile user query into a Python-like plan with loops/branches}

\Statex \Comment{\textbf{Phase 2: Secure Plan Execution}}
\For{instruction $I$ in $\Pi$}
    \Statex \quad \textit{// Step 1: OBSERVE - Gather Environment State}
    \If{$I$ calls $\text{summarize\_screenshot}(query)$}
        \State $O_{raw} \leftarrow \text{Q-VLM}(E.\text{screenshot}, query)$
        \State \textbf{Security Check:} If $\text{Verifier}(O_{raw}, E) = \text{ATTACKED}$ \textbf{abort} 
        \State Update Plan Variables with $O_{raw}$
    \EndIf

    \Statex \quad \textit{// Step 2: VERIFY - Conditional Logic and State Hypothesis}
    \If{$I$ calls $\text{verify\_hypothesis}(observation, hypothesis)$}
        \State $S_{status} \leftarrow \text{Q-VLM}(observation, hypothesis)$
        \State \textbf{Security Check:} If $\text{Verifier}(S_{status}, E) = \text{ATTACKED}$ \textbf{abort} 
        \State Branch Plan Execution based on $S_{status}$ (e.g., if `FAIL' then retry or explore)
    \EndIf

    \Statex \quad \textit{// Step 3: ACT - Find Elements and Execute}
    \If{$I$ calls $\text{find}(desc)$ or $\text{find\_element\_by\_text}(desc)$}
        \State $L_{vis} \leftarrow \text{Q-VLM}(E.\text{screenshot}, desc)$ \Comment{Visual Search}
        \State $L_{dom} \leftarrow \text{Q-VLM}(E.\text{DOM}, desc)$ \Comment{Accessibility Tree Search}
        \State \textbf{Security Check:} If $\text{Verifier}(L_{vis} \cup L_{dom}, E) = \text{ATTACKED}$ \textbf{abort} 
        \If{$L_{vis} \neq \emptyset \lor L_{dom} \neq \emptyset$}
            \State Execute Action (Click/Type) on $E$ at coordinates
            \State \textbf{Verify Action:} Check if $E.\text{screenshot}$ changed 
        \Else
            \State Trigger Retry Loop or Fallback Strategy 
        \EndIf
    \EndIf

    \If{$I$ calls $\text{check\_done}()$}
        \If{$\text{Q-VLM}(E, T) = \text{True}$} \Return \text{Success} \EndIf
    \EndIf
\EndFor

\State \Return \text{Fail} (if plan exhausts without success)
\end{algorithmic}
\end{algorithm}

\subsubsection{Threat Model}
\label{sec:app:threat_model}

\textbf{Attacker Capabilities and Objectives.} We assume an attacker who cannot access or manipulate the P-LLM but can manipulate parts of the execution environment. This aligns with realistic scenarios where attackers control entire malicious websites or inject content within trusted sites (e.g., advertisements, forum posts, product reviews), but do not control all websites the agent interacts with or the underlying agent infrastructure. The attacker embeds malicious visual or textual cues that the Q-VLM perceives during plan execution, with the goal of manipulating the information returned to the P-LLM and thereby steering the agent into adversary-chosen actions.

\textbf{Attacker Knowledge Assumptions.} We assume the \textit{attacker knows the defender's task}, \textit{knows the functions the defender creates for the P-LLM to use}, and \textit{can predict approximately the plans that the defender's P-LLM generates}. These strong knowledge assumptions represent a worst-case scenario that only strengthens our results: any defense effective against such a well-informed attacker will be even more effective against attackers with less information. Under these assumptions, the attacker \textbf{can construct targeted interventions in the environment to steer plan execution along specific branches}. Any function that retrieves information from the environment serves as a potential attack vector, with attack surface expanding as plan complexity increases.

\textbf{Assumption 1: Task prediction.} Task prediction is feasible when attackers target specific websites they own or where they can place advertisements. Each website supports a finite set of user tasks, and attackers may even have access to analytics showing common navigation patterns.

\textbf{Assumption 2: Function prediction.} Function prediction is plausible within known task domains; for CUAs, the space of useful environment information retrieval functions is limited, and tool call functions (e.g., click, type) follow relatively standardized conventions.

\textbf{Assumption 3: Plan pattern prediction.} Plan pattern prediction is supported by our evaluation showing that many planners exhibit similar problem-solving patterns, which become more predictable with descriptive system prompts that provide explicit planning guidance.

\subsubsection{From adaptive to single shot planning}
When we adapt CUAs to work with the Dual-LLM setup, roughly we need to separate perception from action, reasoning, and memory. Thus, we are forced to use a \textbf{hierarchical agent framework setup} rather than only an end-to-end model. All reasoning and action planning is done by the P-LLM. An end-to-end trained CUA model forms the base for our \textbf{perception module} (the Q-VLM), which may be used only to retrieve information to inform conditional logic in a plan or argument values for tool calls. To enhance predictability which can work in favor for utility as well as security, we expand the usual \textbf{action space} of GUI tool calls, with pre-defined tool calls with which the Q-VLM can be called. Specifically, we have one function that summarizes the screenshot within a limited amount of text (\textit{summarize\_screenshot\_content}) based on a short description of what is being looked for. Another function finds the UI elements matching a given text description and returns UI coordinates as floats (\textit{find}). A final function checks whether a task described by a short text has finished and returns a boolean (\textit{check\_done}). Further enhancement to the perception module is given by specific tool calls that parse the DOM and optionally find elements in it with the help of another Q-VLM. Some of these functions have the same functionality as a Q-VLM tool, so they are redundant but in this instance we use this for utility rather than security. One function returns up to a certain length of the text of the elements in the accessibility tree (\textit{get\_page\_elements}), another returns up to a certain length of the text content visible in the current page (\textit{get\_page\_text}). Then, we implement another find function but using the elements from the accessibility tree query a Q-LLM for the element matching the text description given (\textit{find\_element\_by\_text}). Lastly, we add one function (\textit{verify\_hypothesis}) which takes a text describing a hypothesis of a state in which the environment could be in and a text with the actual observation and uses the Q-LLM to compare the two described states and outputs either OK, FAIL or UNKNOWN. 

For \textbf{reasoning}, the P-LLM may use pre-trained knowledge as well as knowledge given via a general system prompt to then plan the full task with all possible states denoted through steps, branches, and loops in one-shot. The difficulty of maintaining utility with Dual-LLM is in not being able to update the plan depending on environment information and iterate and reflect on failures. Many end-to-end models are trained on multi-turn sequential planning setups, and even in most agentic frameworks subtask completion influences and changes the next subtasks to attempt. Needing to predict realistic possible paths for all sequential subtasks is a problem that explodes exponentially in numbers of possible branches and may at first glance seem infeasible to most researchers working on CUAs. Furthermore, short term \textbf{memory} only exists in the form of variables in a program, while long term memory is more similar to other CUAs and can consist of knowledge from model training as well as system prompt.

\subsubsection{Practicality of how to improve utility of single shot planning}
\label{sec:methodology:subsec:dual-llm_cua:subsubsec:practicality}
As we tradeoff a lot of reasoning by design through the restriction of planning the whole agent execution single shot, we share the practical insights that we gained when building the implementation of Dual-LLM for CUAs that improved our utility and made our setup more practical.

\textbf{CUA is not enough, let's add the DOM back in.} Many CUAs do not utilize the DOM at all due to limitations depending on application and OS and the difficulty of handling its verbosity and also parsing a screenshot at the same time. However, we find that the DOM can be a very valuable second source of information that can be easily added into single shot planning and can even enhance the utility of plans. Getting the page elements or the page text can provide a less biased view than a summary of the screenshot from the end-to-end CUA model but may at times be too verbose. For finding elements the DOM can be more precise and when enhanced with an LLM that receives a query for a specific element and all DOM elements it can for example also expose smaller elements less visually noticeable. However, the end-to-end trained CUA can be more dynamic and return related elements even if not an exact match, as it was trained to reason about the query it was given in the context of a user interface. While this could be beneficial for utility if the jump is small enough, it can become a problem when this causes an action that the plan did not foresee or foresaw only later in the plan. 

\textbf{Adding state verification.} To ensure that the CUA is not jumping ahead and outputting a move that has nothing to do with the query, we add a check after the \textit{find} function on whether the verbalized thoughts of the CUA fit with the initial input query to the find function. This is implemented through a call to another LLM with a specific query prompt which asks it to find any discrepancy between instruction and CUA output. If a discrepancy is found the coordinates are not returned and rather `FAIL' is returned. Furthermore, we also add a verification after any action taken on whether the screenshot of the computer changed before or after. If there is no change in state the knowledge that a subtask has not finished can be used further in the plan.

\textbf{Developing general process methodologies (OBSERVE, VERIFY, ACT).} When trying to optimize plans to succeed on CUA tasks, we noticed that especially the first few steps need to be similar in many tasks. First, as the tasks often are only composed of one sentence that do not describe anything about how the computer state is at the beginning of the task, the CUA needs to understand where it is and where it needs to be to succeed in the task. If for example the task were to be ``\textit{Find the Monthly forecast for Manchester, GB for this month}'', then first the plan needs to check whether the browser is open already, if the browser is open it needs to check whether a specific website related to weather is already open. To systematically approach this in a plan we teach the P-LLM in the system prompt to always follow a three step process of observing verifying and then acting. First, for observing, functions such as \textit{summarize\_screenshot\_content}, \textit{get\_page\_elements}, or \textit{get\_page\_text} can be used. Second, the text output of such functions can be used to verify whether the hypothesis of which state the computer is on is correct or not, e.g., does the description for \textit{summarize\_screenshot\_content} fit the hypothesis we are on a website. Third, now depending on whether the hypothesis matches or not the conditional can try to use \textit{find} or \textit{find\_element\_by\_text} to either find a search bar element to find a website to navigate to or if already on a website continue to explore its elements to find the monthly forecast of Manchester. This pattern is a versatile pattern that is useful in many substeps of a plan and works as long as the planner can predict all the different state hypotheses the computer could be in, on the way to finishing a task. This includes many conditional branches and adding fall back options, for when the state is in a different state from the main predicted path or second or third best predicted path.

\textbf{Finding, branching, and looping.} In order to make finding elements as smooth as possible, we add into the prompt how both finding functions can be used and underline that checking whether a coordinate was actually found and whether the action actually changed after the action is taken is important to move on to the next part of the plan. Since elements might not always be found at first try because of the way the query is set by the planner, we prompt the planner to be more descriptive than just one or two words when looking for UI elements. Additionally, we recommend to use both finding functions sequentially, in case one works better than the other for a specific prompt. Finally, we recommend iteration over multiple possible prompts for finding elements needed for the plan in a for loop, which always tries with one find function and then the other, verifies states each time and if coordinates are found the clicking is still also located in the for loop, such that the for loop can be continued in case the clicking fails. Since CaMeL does not have any branches, we recommend the usage of a \textit{no\_op} function to iterate on in the loop if a clicking task has finished early. This does not apply to Fides, as the plan is executed dynamically. 

\textbf{Cookie handling for browser tasks.} When browsing the web, many websites will first have a popup about cookies or privacy consent. The buttons on the popup are not always the same, so we add into the prompt to try to find the correct button with different phrasings of the cookie popup and do iteration as described in above paragraph.

\textbf{A general lesson on how to navigate websites and applications.}
Finally navigation of websites and applications are the least generalizable part, however, we still add advice on how one could generally go about navigating a website, and strongly advise to explore a website before going back into the general search bar and leaving a website or application (which the P-LLM suggests way too often). Some of the strategies include finding an internal search bar or looking for links that may be related to what is being looked for. For example, for the task ``\textit{Browse the natural products database.}'', the task starts out on a website called ``drugs.com''. Strong general prompting to stay on a website that is open is then needed for the planner to not go on general search and finding a different website with natural products that it knows from training like the COCONUT Natural Products Database or the ATLAS Natural Products database. Then after using the search bar in drugs.com to look for the keyword `natural products', the links in the site search do not directly have any titles with the keyword. Hence, general prompting is necessary to advise that content previews of links should also be considered to understand whether they may get you to the right place. The drugs \& medications link is the correct one to navigate to in this instance. On this page now, there are links to browse drugs A-Z, Browse drugs by category and Popular Drug Searches. And finally under Browse drugs by category there is a link to Natural Products. A general lesson from understanding such a benchmark browsing task is that websites are built in such a complex way that browsing them is non-trivial and even humans might need many steps of iteration on failure to get to the right page. It is no wonder then that it is hard for a P-LLM to predict all of the clicks and searches that need to be done to land on the right page. The depth of steps to navigate an environment is not always clear and predictable in advance. We only add some more general advice such as how to scroll a page and systematically recheck if a relevant elements have appeared, to think out of the box and in meta categories and sub categories when prompting to find elements, how menu bars could look like, and to assume that navigation requires going through wider categories and closing into sub categories of the actual element needed.

\subsubsection{Ablation: How to make Fides work}
\label{app:sec:fides}

In Fides, the planner is called iteratively to generate each next action to be executed on the environment. Since outputs from those functions are redacted as variables, the planner can never see any information about the environment. This framework is then different from the CaMeL-style one-shot plan prediction in that it does not enable direct conditionals on the variables output by previous function calls. For instance, the boolean output by the \textit{verify\_hypothesis} function is fully redacted, and thus the planner can not go on a different branch of execution depending on the response from this function. This core difference limits the execution to a simple static plan of execution, which we earlier indicated has very low utility on CUAs. To circumvent it, we choose to relax the process of variable redaction under specified conditions, and we call this new paradigm \textit{Fides-NOVA}. %

\textbf{We relax Fides for CUA by allowing the planner to check boolean variables from our environment functions.} The main potential security implication for this relaxation is that it enables the planner to ultimately get all environment information through infinite repeated tools calls on the same redacted variable. This is a security breach in particular because the planner executes actions iteratively, and thus execution termination cannot be guaranteed comparatively to a fixed-length plan like output by the CaMeL planner. We add several arguments to ensure minimal environment information gain (similar to those necessary in CaMeL for sufficient utility). Those arguments from \textit{Fides-NOVA} are: \textit{max\_steps}, the maximum number of GUI tool calls that are not Q-VLM functions (like \textit{summarize\_screenshot\_content} for instance), and \textit{max\_turn}, the maximum number of any function calls. 
Additionally, we checked that no single variable output from a tool call was reused more than a set \textit{max\_variable\_reuse} times. These three arguments enable both forced task termination and limited environment information gain from the planner. In practice \textit{max\_steps}=15, \textit{max\_turn}=70, and \textit{max\_variable\_reuse}=5.

\section{Further Details on Redundancy Defenses}
\label{sec:app:redundancy_details}

This section provides technical details on the redundancy-based verifier modules introduced in \cref{sec:residual_attack_surface}.

\subsection{Verifier Module Architecture}

We treat functions that call the Q-VLM as potentially faulty nodes since these receive information from the environment and can influence plan execution. For each such function, we introduce a \textbf{verifier} module that cross-checks outputs before they populate plan variables. The verifiers use either second sources of information (e.g., DOM alongside screenshot) or independently-instantiated models to catch outputs that a single Q-VLM might produce under adversarial manipulation. If any verification check detects a potential attack, execution halts and the trace is returned to the user.

The verifiers rely on the empirically observed low transferability of adversarial examples between diverse models \citep{schaeffer2024failurestransferableimagejailbreaks}: optimization-based attacks targeting the Q-VLM are less likely to fool a distinct verifier simultaneously. This remains a best-effort defense: since verifiers are probabilistic models rather than architecturally isolated components, they offer no formal guarantees beyond the core Dual-LLM control-flow protection.

\subsection{DOM Consistency}

The verifier uses an independent VLM to check whether the Q-VLM's visual findings align with the ground-truth DOM structure. Specifically, the verifier receives the DOM alongside the proposed action (coordinates from \texttt{find}) and the original instruction, and flags mismatches that could indicate hallucinations or clicks on disguised elements such as transparent overlays or elements in ad sections. The DOM provides tagged subsections (e.g., ``Advertisement'' frames in static Google banners) that can reveal mismatches not visible in the screenshot alone. However, as we show in \cref{app:cookie_attack_details}, attackers who control HTML5 ad banners can rewrite the DOM to bypass these frame-level checks.

\subsection{Multi-Modal Consensus}

The verifier (a second independent VLM, typically from a different vendor than the Q-VLM) receives both the screenshot and the Q-VLM's output alongside the original instruction. Its task is to detect visual anomalies or semantic mismatches between the Q-VLM's response and what the screenshot shows—for example, recognizing that an element labeled ``Login'' in the Q-VLM's output is actually a pixel-injected advertisement, or that raw HTML camouflaged as legitimate in the DOM appears malicious visually. By employing a distinct verifier model (e.g., a closed-source VLM different from an open-source Q-VLM), optimization-based attacks targeting the Q-VLM are less likely to simultaneously fool the verifier. As we show in \cref{app:pixel_attack_details}, however, this defense can still be circumvented when adversaries optimize perturbations to produce plausible Q-VLM outputs that align with both verifier inputs.

\section{Further Details on Attack Implementations}
\label{sec:app:further_details_attack}

This section provides technical details on our attack implementations.

\subsection{Positioning Relative to Prior Work}
\label{app:attack_positioning}

Similar to \citet{zhang2025attacking}, we embed our attack into seemingly legitimate UI pop-ups, but we extend the realism by specifically targeting ad banners as vectors for fake pop-ups, thereby mimicking the realistic threat models proposed by \citet{aichberger2025mip,evtimov2025wasp}. These attacks differ from other potential OS-level entry points, such as those presented in \citet{chen2025obvious} which insert malicious links or toggle wraps directly into the UI, or the website prompt injection attacks in \citet{kuntz2025osharm} which embed adversarial text directly into website layouts. Our pixel-based attack can be viewed as a specific instantiation of the image-based attacks presented in \citet{aichberger2025mip}, as our adversarial images are optimized toward a single objective and initialized from a simple white background rather than from synthetic images.

\subsection{Cookie Attack Technical Details}
\label{app:cookie_attack_details}

\subsubsection{Attack Mechanism}

The attack exploits a fundamental design requirement of defended CUAs: agents must handle cookie popups and advertisements before interacting with websites. An example Python code snippet demonstrating this requirement is provided in \cref{app:cookie_snippet}. The attack generalizes to any predictable utility routine that must appear early in task execution, such as cookie handling, login verification, or welcome screen dismissal.

\begin{figure*}[t]
\centering

\begin{tikzpicture}[node distance=0.6cm and 0.6cm, every node/.style={inner sep=0pt}]

\node (img1) {\includegraphics[width=0.48\textwidth]{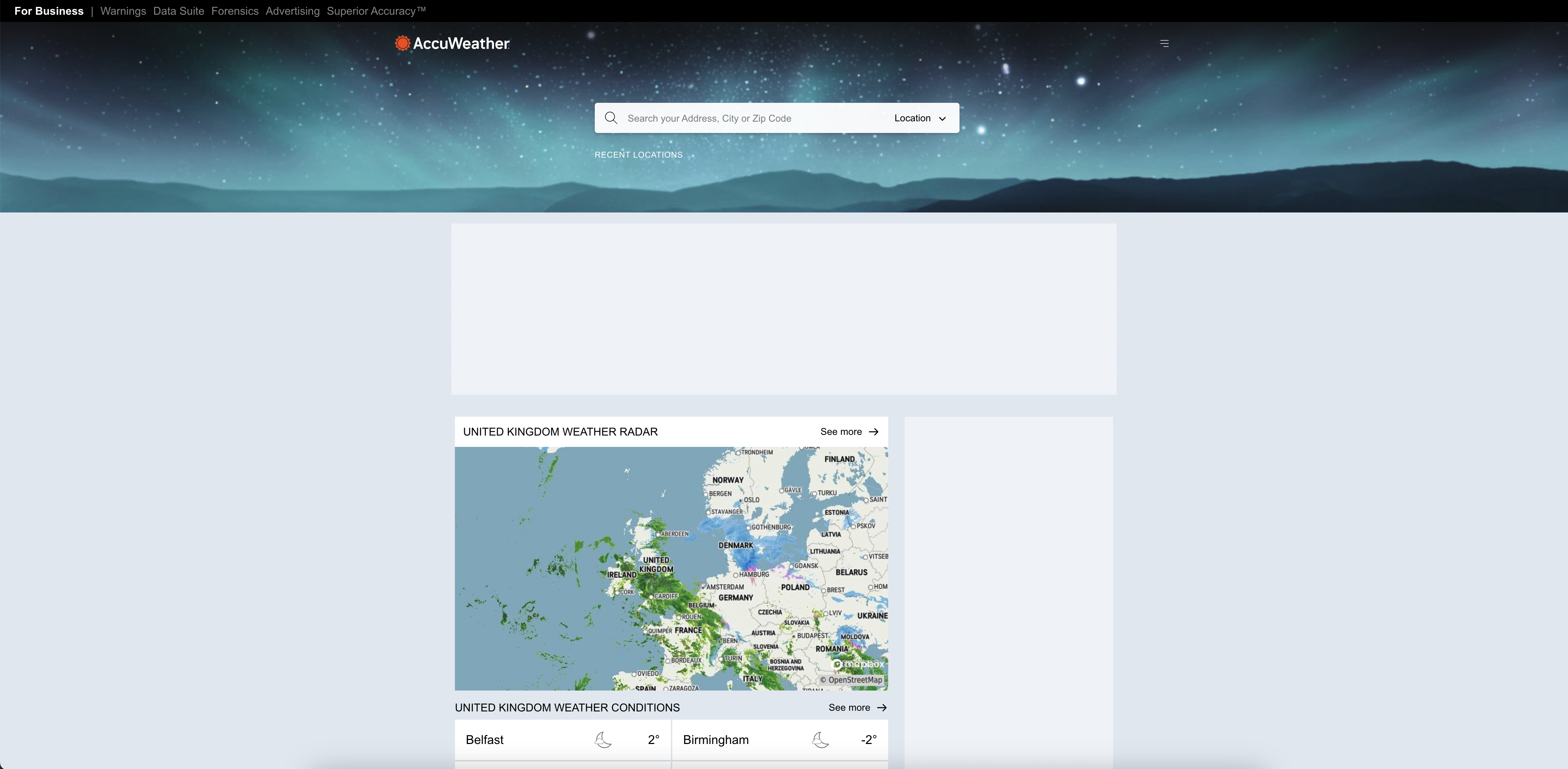}};
\node (img2) [right=of img1] {\includegraphics[width=0.48\textwidth]{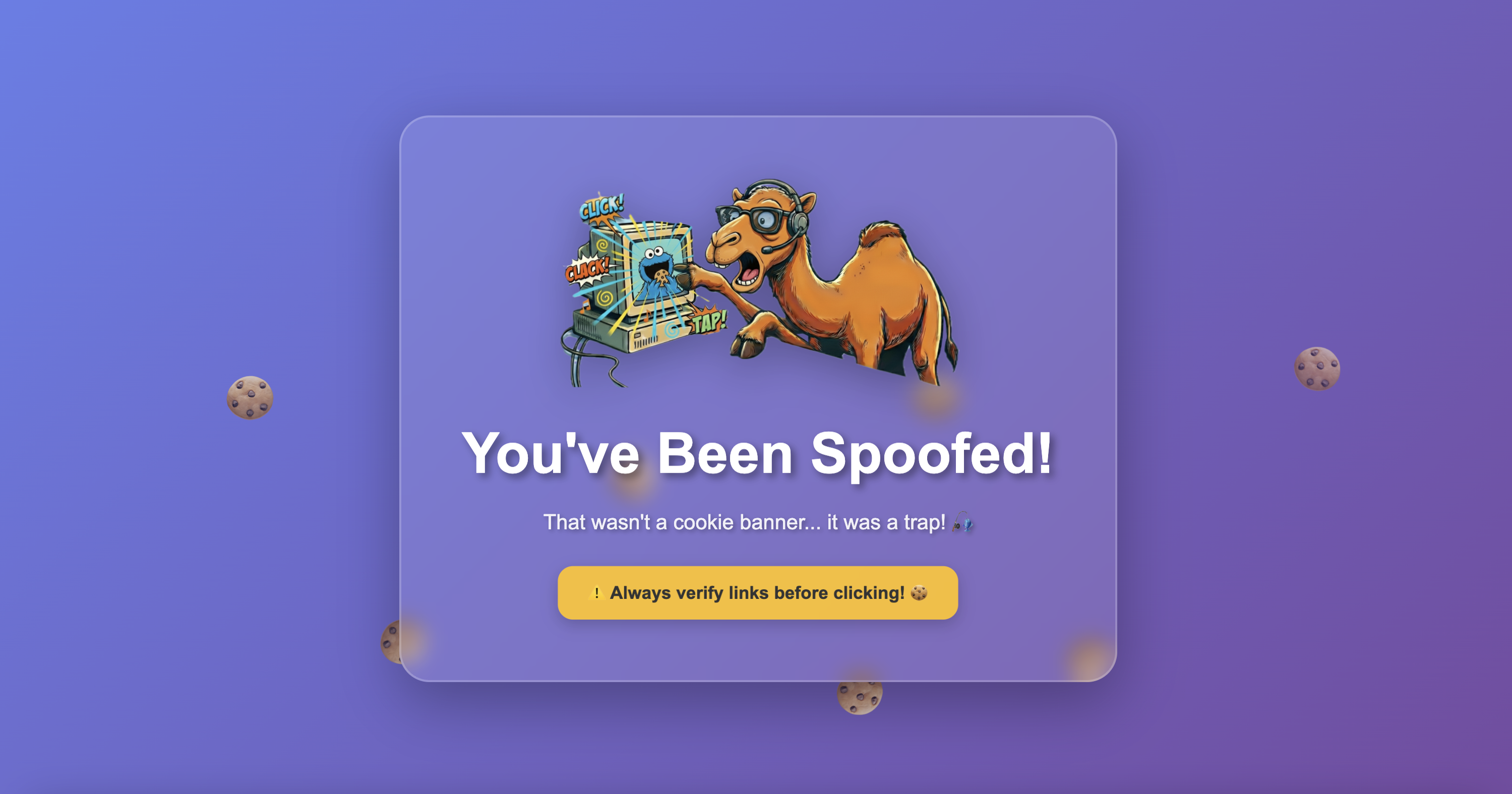}};
\node (img3) [below=of img1] {\includegraphics[width=0.48\textwidth]{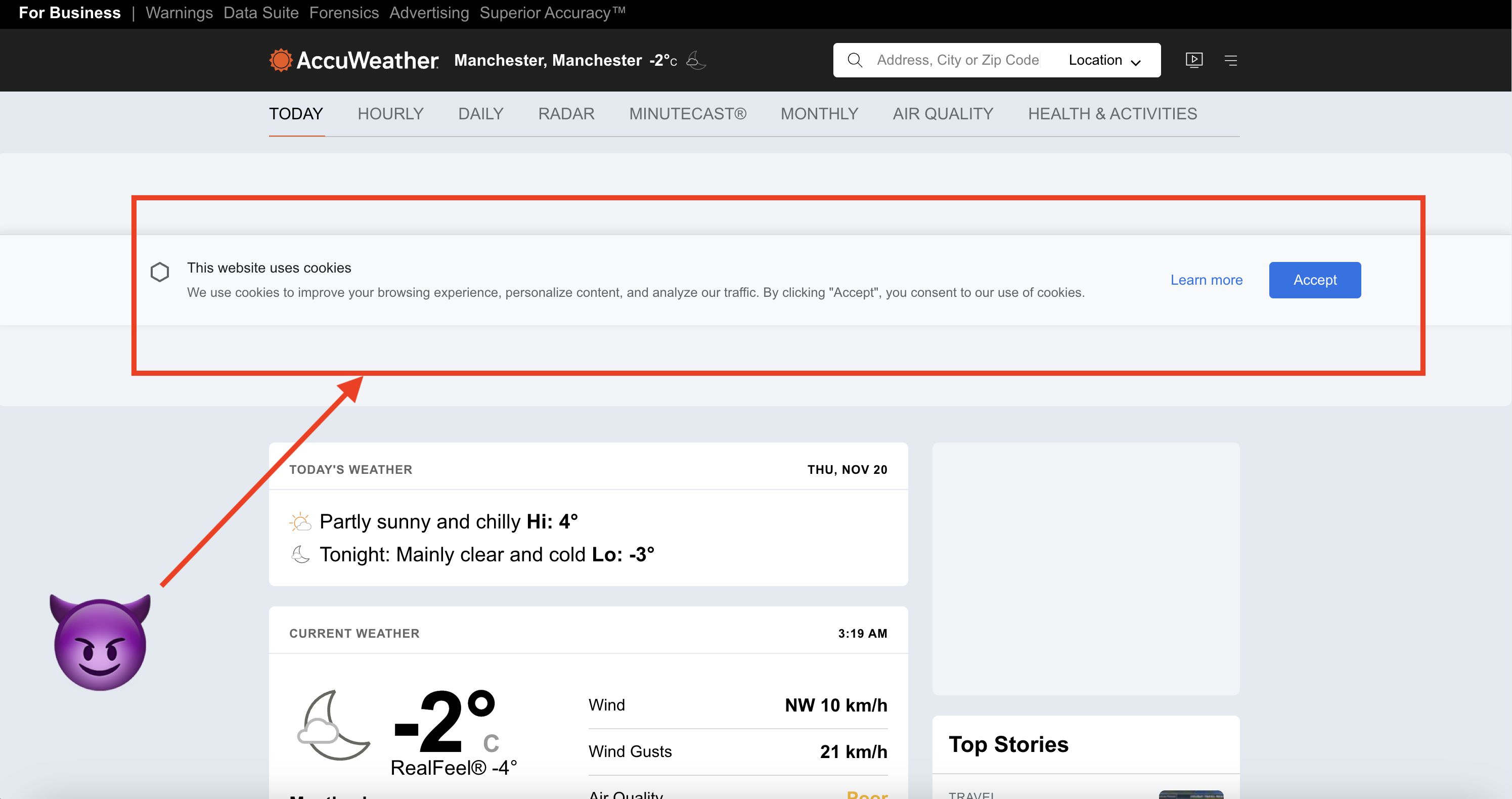}};
\node (img4) [below=of img2] {\includegraphics[width=0.48\textwidth]{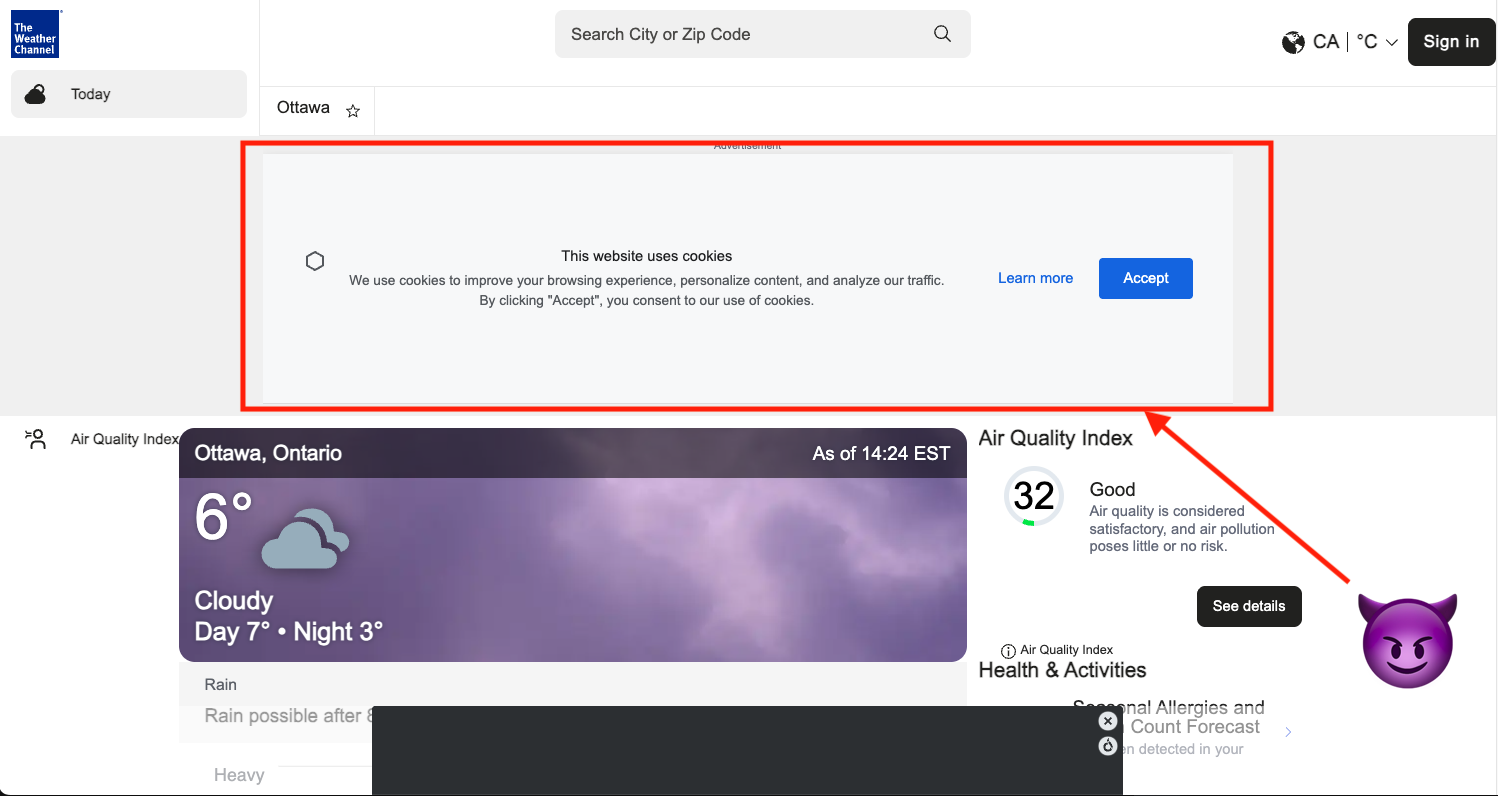}};

\node[draw, circle, fill=white, inner sep=1pt, font=\bfseries, anchor=south east] at (img1.south east) {1};
\node[draw, circle, fill=white, inner sep=1pt, font=\bfseries, anchor=south east] at (img2.south east) {4};
\node[draw, circle, fill=white, inner sep=1pt, font=\bfseries, anchor=south east] at (img3.south east) {2};
\node[draw, circle, fill=white, inner sep=1pt, font=\bfseries, anchor=south east] at (img4.south east) {3};

\draw[-{Latex[length=3mm]}] (img4.north) -- (img2.south); %
\draw[-{Latex[length=3mm]}] (img1.south) -- (img3.north); %
\draw[-{Latex[length=3mm]}] (img3.east) -- (img4.west);   %

\end{tikzpicture}

\caption{\textbf{Cookie attack execution flow.}
Illustration of a \textit{multi-step} and \textit{long-range} cookie attack:
(1) legitimate homepage, (2) spoofed intermediate page, (3) trusted redirect, (4) final malicious landing page.}
\label{fig:cookie_attack}
\end{figure*}

We embed fake cookie popups within ad banners. When the \texttt{find} function is called with instructions to locate cookie popups, the Q-VLM returns coordinates for the fabricated popup, redirecting the agent to an attacker-controlled webpage (\textit{spoofed.html}, shown in \cref{fig:cookie_attack}, top right). While other attacks such as denial of service are possible, we consider redirection to attacker-controlled websites the most harmful outcome under this threat model. We evaluate attacks on two widely adopted advertisement formats: (1) static Google advertisement banners, which allow advertisers to specify a hyperlink and image, and (2) embedded HTML5 advertisement banners, which support full HTML snippets and thus provide a larger attack surface. While our concrete implementation targets Chrome-based tasks in GDPR jurisdictions where cookie popups are mandatory, the underlying vulnerability generalizes to any predictable UI interaction the adversary can anticipate and control.

\subsubsection{Defense Level Interaction}

\textbf{DOM Consistency.} An additional Q-VLM receives the DOM alongside the proposed action (coordinates from \texttt{find}) and the original instruction. For static Google ad banners, the standardized HTML code features a common ``Advertisement'' frame containing placeholders for an image and hyperlink (the only elements accessible to the attacker). The DOM provided to the Q-VLM includes this ``Advertisement'' string and the coordinates of the ad banner, successfully flagging any click attempts within this region. However, HTML5 ads present a vulnerability: the DOM communicates only the leaf nodes of the webpage's HTML architecture. When a fabricated cookie popup is embedded within the ad banner, it replaces the ``Advertisement'' frame with a cookie frame in the DOM, effectively circumventing DOM-based defenses.

\textbf{Multi-Modal Consensus.} An additional independent Q-VLM call validates the screenshot alongside function outputs and the original instruction. Although fabricated cookie banners are clearly visible within ad banners to human observers, enabling clear distinction for the screenshot-based verifier requires considerable effort. Achieving effective attack detection without degrading benign task utility is challenging, consistent with the weak success rates observed for prompt-based defenses \citep{evtimov2025wasp}. Even with an optimal tradeoff, we show in \cref{app:pixel_attack_details} that this defense can be circumvented via optimized pixel attacks.

\subsubsection{Multi-Step Attack with Hop Website}
\label{app:cookie_attack_multistep}

The one-step attack assumes an adversary can directly specify the \textit{spoofed.html} hyperlink in the victim's homepage ad banner, which in practice is usually blocked by hyperlink verification on ad-hosting sites. We bypass this by routing the attack through an intermediary benign ``hop'' website that does not verify its banner hyperlinks: a first fabricated cookie popup redirects to the hop site, and a subsequent popup there redirects to \textit{spoofed.html}.

This attack succeeds when the benign task requires visiting more than two webpages: the plan must then anticipate handling at least two cookie popups, enabling our attack. The multi-step scheme generalizes to any number of webpage visits that the planner anticipates for task completion. The underlying vulnerability is that each \texttt{find(``cookie popup'')} call in the plan resolves independently against whatever is currently on screen, without context for which plan step the agent has reached. The planner cannot encode such context because it never sees page identity at runtime.

\textbf{Long-range variant.} We further extend the attack's potential harm by introducing a \textit{long-range} effect. Instead of embedding the first fabricated cookie popup on the initial webpage, we place it on a later webpage that the adversary predicts the agent will visit based on the accessible plan. This approach increases stealth by delaying the malicious redirection until deeper into task execution. The complete multi-step long-range attack is illustrated in \cref{fig:cookie_attack}.

\subsection{Pixel Attack Technical Details}
\label{app:pixel_attack_details}

\subsubsection{Objective and Rationale}

We demonstrate the brittleness of redundancy defenses against data-flow attacks. Even at our strongest redundancy defense configured to heavily prioritize security over utility (accepting a high false positive rate), adversaries can craft attacks that evade both screenshot-based and DOM-enhanced Q-VLM verification. The banner is optimized with an objective function that induces malicious Q-VLM outputs when querying \texttt{find}. Returning the ad banner's coordinates would be immediately flagged by the redundancy defense; we instead optimize the Q-VLM to return \textit{plausible} thought traces and coordinates that align with the element search instruction. This ensures the defense does not flag the attack on either front: the screenshot and DOM appear benign (containing only seemingly random pixel perturbations), while the Q-VLM's returned action and thought trace appear legitimate and consistent with the task. Such attacks could be exploited by advertisers, such as pharmaceutical companies, to artificially increase traffic to their product pages by redirecting CUA agent traffic on commercial websites.

\subsubsection{Implementation}

\begin{figure}[htbp]
\centering
\includegraphics[width=\linewidth]{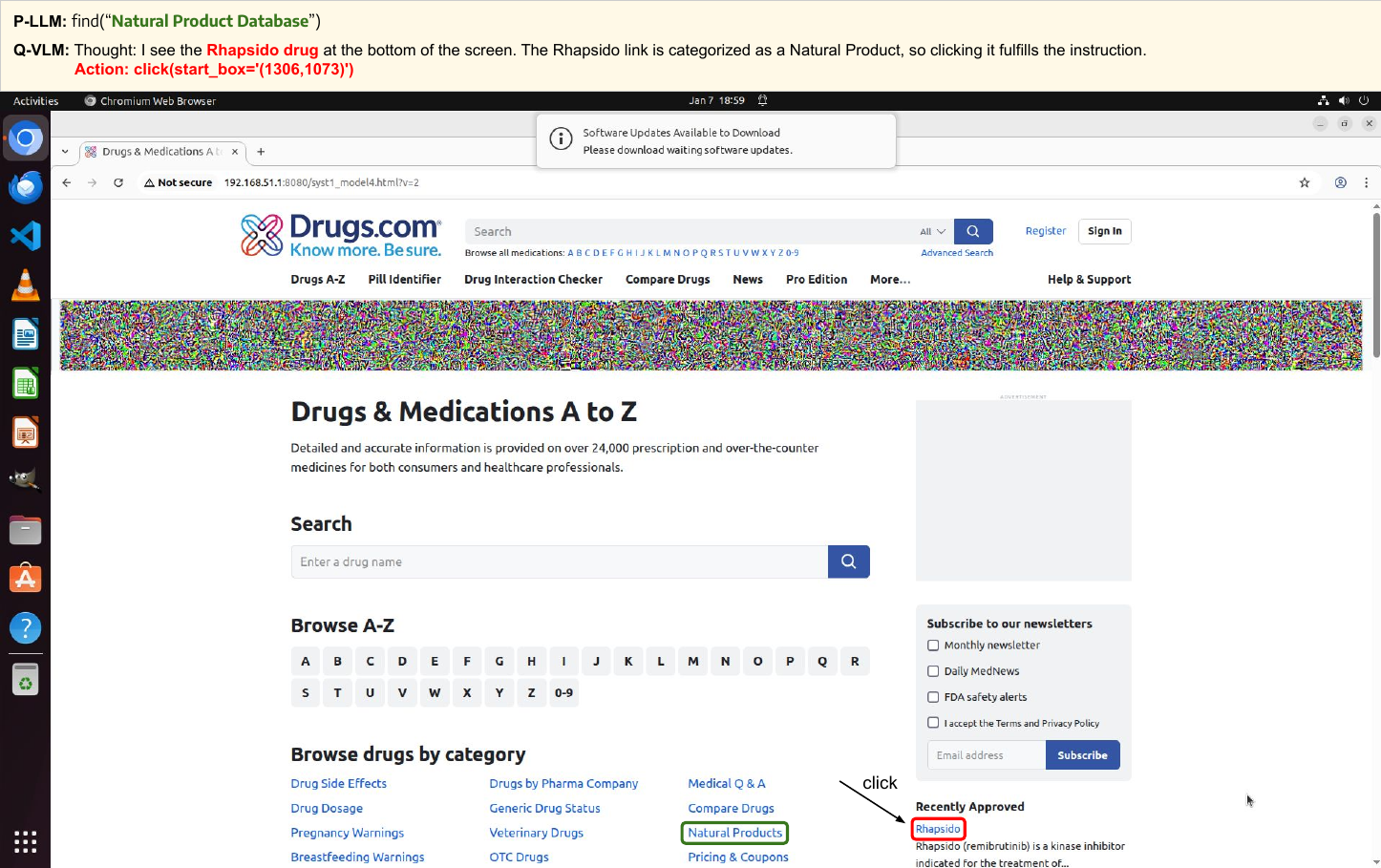}
\caption{An example of how the pixel attack looks like. Q-VLM is tricked into clicking at the specific drug instead of the ``Natural Product Database'', and to provides justifiable explanation (that the drug is categorized as natural product) hence the action is not flagged by the redundancy defense (Level 2).}
\label{fig:pixel_attack}
\end{figure}

We perform gradient-based optimization of a masked image region to lead UI-TARS to generate attacker-chosen token sequences. Concretely, we optimize a banner region at the top of a \textit{www.drugs.com} screenshot to make the model click on a specific drug link (``Rhapsido'') instead of the intended target (``Natural Product Database''), while generating a justification that the clicked drug is categorized as a natural product. The target sequence is shown in \cref{fig:pixel_attack}.

The attack optimizes pixel values in the model's patch embedding space using backpropagation through the vision encoder, with straight-through gradient estimation to handle PNG quantization during image processing. We optimize only patches where all pixels fall within the mask, keeping non-masked patches fixed at their base screenshot values.

Robustness is achieved through Expectation Over Transformations (EOT) by averaging gradients over 2 augmented samples per iteration, applying Gaussian noise with $\sigma=0.005$ and extra noise in the top-center region (clock location) with $\sigma=0.05$. We use the Adam optimizer with learning rate 0.8, cosine annealing learning rate schedule (200-iteration warm restarts until iteration 600, then smooth decay to $\text{lr}_{\min}=8 \times 10^{-4}$), gradient clipping at $L_2$ norm = 10, and run for 2000 iterations on UI-TARS-1.5-7B.

\subsubsection{Extensions and Future Work}

We leave potential multi-objective and multi-Q-VLM pixel attacks to future work, as \citet{aichberger2025mip} have already demonstrated techniques for constructing such attacks. We also defer multi-step pixel attacks, noting that an $n$-step attack would simply require chaining $n$ instances of the one-step attack described above.

\section{Further Utility Measurements on OSWorld}
\subsection{Base Utility on OSWorld}
\label{sec:app:base_utility}
\begin{table}[htbp]
\scriptsize
\centering
\caption{\textbf{OSWorld task distribution}. Claude Sonnet 4.5, OpenCUA-32B, and UITars-1.5-7B denote the successful tasks (excluding those that are denoted infeasible to evaluate) for these models in each category when restricting the models to at most 15 steps, a history of 15 and a maximal output token length for the VLM of 4096 (pass@1). Only for OpenCUA we cut off the history at 5 due to slow throughput. To calculate the overall success rate for these three models, we add back the 30 tasks that are defined as not automatically evaluatable and thus by default set to success in the OSWorld benchmark. While our success rates for OpenCUA and UITars are similar to those denoted in the OSWorld benchmark, the success rate for Claude is significantly lower, which we attribute to the much lower step size that we chose (15 vs. 50).}
\label{tab:task_distribution}
\begin{tabular}{lrrrrr}
\toprule
\textbf{Category} & \textbf{All} & \textbf{Not Infeas.} & \textbf{Claude} & \textbf{OpenCUA} & \textbf{UITars} \\
\midrule
Chrome & 46 & 43 & 14 &13 & 11 \\
GIMP & 26 & 16 & 11 & 9& 6 \\
LibreOffice Calc & 47 & 46 & 13 & 3 & 3 \\
LibreOffice Impress & 47 & 47 & 17 &14 & 9 \\
LibreOffice Writer & 23 & 22 & 12 &7 & 8 \\
Multi-apps & 101 & 100 & 7 &3 & 3 \\
OS & 24 & 19 & 11 &7 & 4 \\
Thunderbird & 15 & 14 & 8 &7 & 4 \\
VLC & 17 & 14 & 4 &4 & 2 \\
VS Code & 23 & 18 & 12 & 9& 8 \\
\midrule
\textbf{Total} & \textbf{369} & \textbf{339} & \textbf{109} & \textbf{76}& \textbf{60} \\
\midrule
\textbf{Success Rate (\%)} & & & \textbf{37.67} &\textbf{28.72} & \textbf{24.39} \\
\bottomrule
\end{tabular}
\end{table}

The base utility on OSWorld per task is shown in \cref{tab:task_distribution}. Of the 369 total tasks, 30 are marked as automatically unmeasurable, and marked as automatically successful by the benchmark. We adhere to this measurement in our table. For reference, published OSWorld benchmark numbers are 24.5$\pm$1.2\% with 15 steps at pass@2 for UITars-1.5-7B, 29.7$\pm$1.1\% with 15 steps at pass@3 for OpenCUA-32B, and 58.1\% for Claude Sonnet 4.5 with a higher step count of 50 steps.

\subsection{Plan Analysis Methodology}
\label{app:sec:plan_analysis_methodology}
To quantify the structural differences between plans produced for CUA tasks and those produced for typed-API benchmarks, we build a static analysis pipeline that extracts directed acyclic graphs (DAGs) from the Python-like plans emitted by the P-LLM. Each plan is parsed into an abstract syntax tree (AST), and every tool call (e.g., \texttt{find}, \texttt{verify\_hypothesis}, \texttt{left\_single}) becomes a node in the graph. We construct two types of edges: \emph{sequential edges} connect consecutive tool calls within the same basic block, capturing the plan's linear execution order, while \emph{data-flow edges} link a tool call to any downstream call that consumes its output through a shared variable. Variable-to-node bindings are tracked through assignments, including through conditional branches.
Each \texttt{if}-statement in the plan generates a \emph{branch point}. We classify every branch condition into one of four categories by tracing the data-flow origin of the variables that appear in the condition: \emph{semantic verification} branches are gated by \texttt{verify\_hypothesis} or \texttt{check\_done} outputs; \emph{action status guards} check whether a prior UI action (e.g., \texttt{click}, \texttt{type\_text}) succeeded via its \texttt{.status} field; \emph{observation-conditioned} branches depend on Q-VLM perception calls such as \texttt{find} or \texttt{get\_page\_elements}; and \emph{control-flow logic} branches involve only plan-internal values (loop counters, literal flags, list indices) with no transitive dependency on any tool output. 
From the DAG we derive several aggregate metrics. The \emph{tool call count} is the number of graph nodes, \emph{branch count} the number of \texttt{if}-statements, and \emph{LLM-conditioned branches} the subset whose condition depends (directly or transitively) on any Q-VLM output. For intra-task diversity, we generate $k{=}3$ independent plans per task and compute pairwise metrics across them: \emph{sequence edit distance} (normalized Levenshtein distance over tool-call sequences), \emph{cosine similarity} (over frequency vectors of tool names), and \emph{pattern Jaccard similarity} (over the set of distinct execution paths, where each path is a complete root-to-leaf tool-call sequence through the DAG). These diversity metrics characterize whether additional plan samples explore meaningfully different strategies or merely paraphrase the same approach. We report per-plan means $\pm$ 95\% confidence intervals across $N{=}886$ CaMeL-CUA-NOVA and CaMeL-CUA-unoptimized plans, and $N{=}242$ CaMeL-AgentDojo plans in \cref{tab:plan_stats}, with per-application-suite breakdowns in \cref{tab:plan_stats_per_suite}. For CaMeL-CUA methods, $N{=}886$ is the number of Pass@3 plans generated over the set of OSWorld non-infeasible tasks. For CaMeL-AgentDojo, it is Pass@3 as well, over the full benchmark.

\subsection{Plan statistics details}
\label{app:sec:plan_stats}

\begin{table}[htbp]
\centering
\caption{Plan complexity comparison across different CaMeL adaptations. CaMeL-CUA-NOVA plans on CUA tasks require an order of magnitude more tool calls and branches than CaMeL-AgentDojo plans, reflecting the unbounded state space of OS environments. CaMeL-CUA-unoptimized falls between the two. Values are per-plan means with a $\pm$ 95\% confidence interval. Branch condition categories are percentages of all branching conditions.}
\label{tab:plan_stats}
\small
\begin{tabular}{l c c c}
\toprule
\textbf{Metric} & \textbf{CaMeL-CUA-NOVA} & \textbf{CaMeL-CUA-unoptimized} & \textbf{CaMeL-AgentDojo} \\
 & $N{=}886$ & $N{=}886$ & $N{=}242$ \\
\midrule
\multicolumn{4}{l}{\textit{Plan Structure}} \\
\quad Code Lines                     & $213.3 \pm 7.5$   & $71.6 \pm 3.7$    & $51.8 \pm 6.9$ \\
\quad Tool Calls                     & $41.1 \pm 1.6$    & $19.8 \pm 1.7$    & $4.9 \pm 0.4$ \\
\quad Data-Flow Edges                & $21.4 \pm 1.0$    & $9.1 \pm 3.1$     & $1.2 \pm 0.3$ \\
\quad Sequential Edges               & $130.6 \pm 8.3$   & $30.7 \pm 3.7$    & $4.6 \pm 0.6$ \\
\midrule
\multicolumn{4}{l}{\textit{Branching}} \\
\quad All Branches                   & $39.7 \pm 1.7$    & $11.3 \pm 0.8$    & $3.7 \pm 0.5$ \\
\quad LLM-Conditioned                & $24.2 \pm 1.1$    & $7.9 \pm 0.7$     & $1.9 \pm 0.2$ \\
\midrule
\multicolumn{4}{l}{\textit{Branch Condition Categories (\%)}} \\
\quad Semantic Verification          & 22.9              & 0.0               & 6.3 \\
\quad Action Status Guards           & 27.2              & 4.5               & 0.0 \\
\quad Observation-Conditioned        & 40.5              & 88.4              & 14.4 \\
\quad Control Flow Logic             & 9.4               & 7.0               & 79.4 \\
\midrule
\multicolumn{4}{l}{\textit{Intra-Task Plan Diversity}} \\
\quad Sequence Edit Distance         & 0.435             & 0.568             & 0.239 \\
\quad Cosine Similarity              & 0.837             & 0.346             & 0.875 \\
\quad Jaccard Similarity             & 0.044             & 0.001             & 0.393 \\
\bottomrule
\end{tabular}
\end{table}

\begin{table}[htbp]
\centering
\caption{Per-task suite plan statistics. CaMeL-CUA-NOVA plans on OSWorld are substantially larger and more branched than CaMeL-CUA-unoptimized or CaMeL-AgentDojo across all application suites. Chrome and Thunderbird demand the most complex plans. AgentDojo suites show high Jaccard similarity due to repeating tool patterns across simple tasks.}
\label{tab:plan_stats_per_suite}
\small
\begin{tabular}{l r r r r r r}
\toprule
\textbf{Suite} & \textbf{Tool} & \textbf{All} & \textbf{LLM} & \textbf{Edit} & \textbf{Cosine} & \textbf{Jaccard} \\
               & \textbf{Calls} & \textbf{Br.} & \textbf{Br.} & \textbf{Dist.} & \textbf{Sim.} & \textbf{Sim.} \\
\midrule
\multicolumn{7}{l}{\textit{CaMeL-CUA-NOVA}} \\
Chrome              & 50.8 & 52.3 & 33.7 & 0.394 & 0.920 & 0.038 \\
Thunderbird         & 45.9 & 48.6 & 29.7 & 0.458 & 0.896 & 0.000 \\
Multi Apps          & 47.7 & 41.0 & 25.5 & 0.473 & 0.834 & 0.069 \\
LibreOffice Calc    & 43.6 & 42.3 & 25.8 & 0.473 & 0.804 & 0.029 \\
LibreOffice Impress & 35.7 & 40.3 & 21.5 & 0.379 & 0.931 & 0.054 \\
VS Code             & 36.2 & 31.2 & 19.9 & 0.542 & 0.883 & 0.007 \\
VLC                 & 35.1 & 34.2 & 22.0 & 0.409 & 0.822 & 0.007 \\
GIMP                & 29.4 & 33.0 & 19.9 & 0.473 & 0.865 & 0.013 \\
OS                  & 24.0 & 21.5 & 13.7 & 0.453 & 0.836 & 0.061 \\
LibreOffice Writer  & 19.2 & 22.6 & 12.0 & 0.264 & 0.492 & 0.026 \\
\midrule
\multicolumn{7}{l}{\textit{CaMeL-CUA-unoptimized}} \\
Multi Apps          & 25.5 & 12.7 &  8.4 & 0.584 & 0.421 & 0.000 \\
Thunderbird         & 24.0 & 10.0 &  9.0 & 0.608 & 0.253 & 0.000 \\
LibreOffice Impress & 21.8 & 12.4 & 10.2 & 0.611 & 0.373 & 0.001 \\
VLC                 & 21.6 & 11.9 & 10.0 & 0.688 & 0.364 & 0.000 \\
Chrome              & 18.7 & 11.2 &  9.1 & 0.607 & 0.462 & 0.001 \\
LibreOffice Calc    & 16.0 & 12.7 &  6.8 & 0.496 & 0.230 & 0.000 \\
VS Code             & 13.4 &  6.0 &  4.3 & 0.695 & 0.246 & 0.000 \\
GIMP                & 13.2 &  8.3 &  7.0 & 0.567 & 0.211 & 0.007 \\
OS                  & 11.9 &  6.5 &  5.4 & 0.454 & 0.297 & 0.000 \\
LibreOffice Writer  &  8.9 &  8.7 &  4.2 & 0.393 & 0.174 & 0.005 \\
\midrule
\multicolumn{7}{l}{\textit{CaMeL-AgentDojo}} \\
Travel              &  7.9 &  5.3 &  3.9 & 0.201 & 0.865 & 0.667 \\
Slack               &  5.6 &  4.5 &  2.2 & 0.229 & 0.892 & 0.383 \\
Workspace           &  4.7 &  3.5 &  1.8 & 0.254 & 0.865 & 0.368 \\
Banking             &  3.7 &  2.9 &  1.7 & 0.186 & 0.907 & 0.444 \\
\bottomrule
\end{tabular}
\end{table}

\begin{figure*}[t]
    \centering
    \begin{subfigure}[b]{0.58\textwidth}
        \centering
        \includegraphics[width=\textwidth]{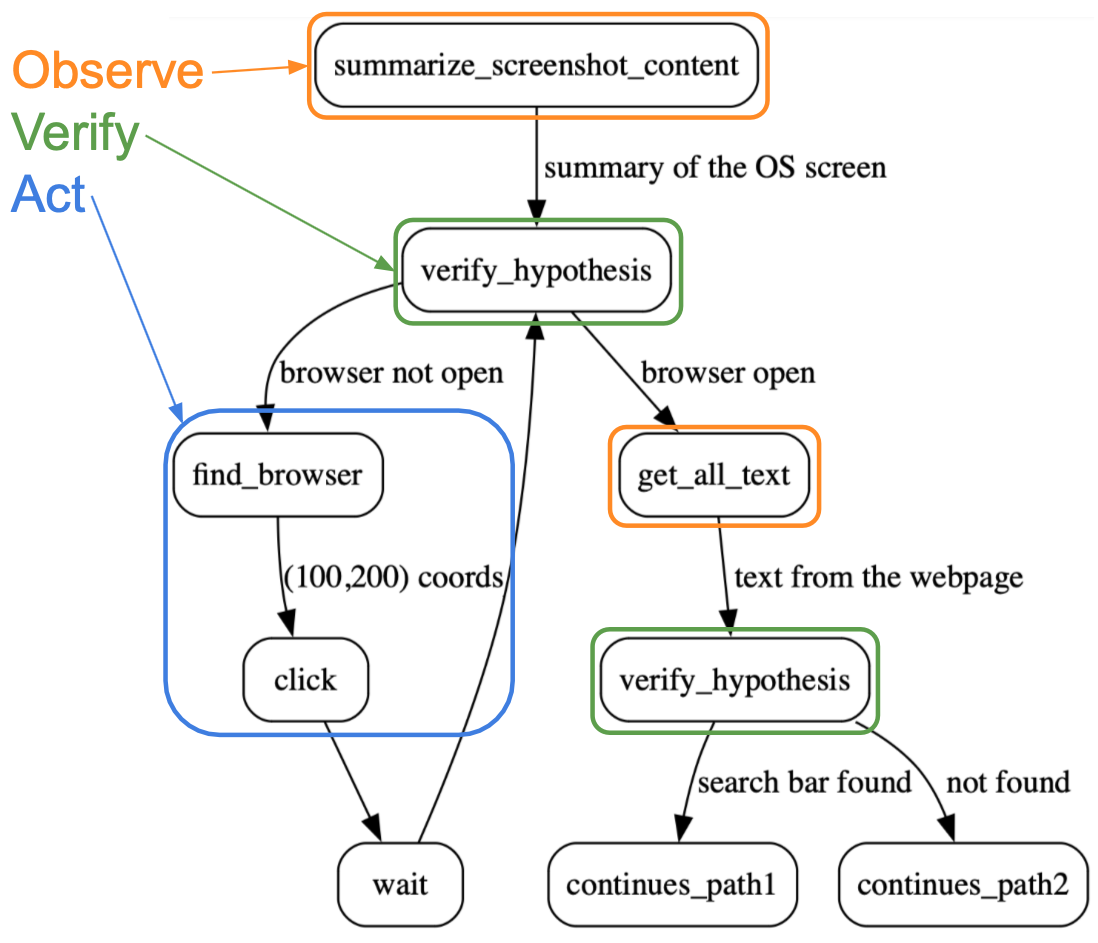}
        \caption{CaMeL-CUA-NOVA. Task: \textit{Find the monthly weather forecast for Manchester, UK}.}
        \label{fig:camel_cua_graph_detailed}
    \end{subfigure}
    \hfill
    \begin{subfigure}[b]{0.27\textwidth}
        \centering
        \includegraphics[width=\textwidth]{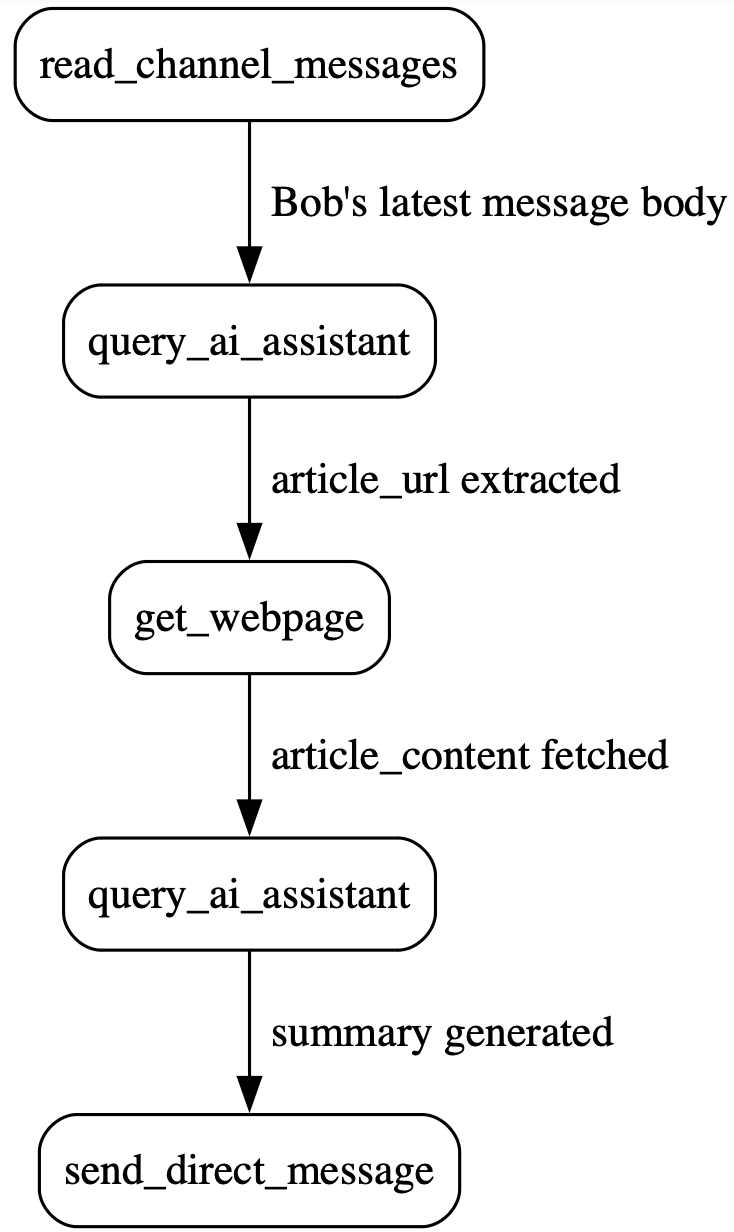}
        \caption{CaMeL-AgentDojo. Task: \textit{Summarize the article that Bob posted in 'general' channel and send it to Alice}.}
        \label{fig:standard_cua}
    \end{subfigure}
    \caption{Side-by-side comparison of our Observe-Verify-Act methodology for planning CUA tasks on the \textit{OSWorld} benchmark and a selected task on the \textit{AgentDojo} benchmark.}
    \label{fig:graph_comparison}
\end{figure*}

\subsection{Statistics on Performance with different Planner Models}
\label{sec:app:planner_stats}
\begin{table*}[htbp]
\centering
\caption{Planner Model Performance on a minimal subset of OSWorld tasks with CaMeL. We pick 8 successful chrome tasks and 1 successful task from each other category from the UITars successful tasks subset, resulting in 17 tasks that we evaluate with 9 different planners using UITars as the Q-VLM. GPT-5 performs by far the best followed by Grok-4.}
\label{tab:planner_model_comparison}
\small
\resizebox{\textwidth}{!}{
\begin{tabular}{lccccccccc}
\toprule{}
\textbf{Metric} & \textbf{GPT-5} & \textbf{Grok-4} & \textbf{GPT-5.1} & \textbf{Gemini 3} & \textbf{Gemini 2.5} & \textbf{Claude} & \textbf{GPT-OSS} & \textbf{Kimi} & \textbf{DeepSeek} \\
 & & & &\textbf{Pro} & \textbf{Pro} & \textbf{Sonnet 4.5} & \textbf{120B} & \textbf{K2} & \textbf{R1-0528} \\
\midrule
\multicolumn{10}{l}{\textit{Success Rates by Application}} \\
Chrome & 7/8 & 4/8 & 2/8 & 1/8 & 1/8 & 1/8 & 3/8 & 2/8 & 2/8 \\
Other Apps & 5/9 & 6/9 & 4/9 & 5/9 & 5/9 & 4/8 & 1/9 & 3/9 & 2/9 \\
\midrule
\multicolumn{10}{l}{\textit{Overall Performance}} \\
\textbf{Overall} & \textbf{12/17} & \textbf{10/17} & \textbf{6/17} & \textbf{6/17} & \textbf{6/17} & \textbf{5/17} & \textbf{5/17} & \textbf{4/17} & \textbf{4/17} \\
\midrule
\multicolumn{10}{l}{\textit{Cumulative Pass@k Performance (\%)}} \\
\textbf{Pass@1} & 17.6 & 29.4 & 17.6 & 29.4 & 23.5 & 23.5 & 6.2 & 11.8 & 0.0 \\
\textbf{Pass@2} & 52.9 & 41.2 & 29.4 & 29.4 & 23.5 & 29.4 & 6.2 & 17.6 & 0.0 \\
\textbf{Pass@3} & 64.7 & 47.1 & 35.3 & 35.3 & 23.5 & 29.4 & 18.8 & 17.6 & 0.0 \\
\textbf{Pass@4} & 64.7 & 52.9 & 35.3 & 35.3 & 23.5 & 29.4 & 18.8 & 23.5 & 17.6 \\
\textbf{Pass@5} & 70.6 & 58.8 & 35.3 & 35.3 & 35.3 & 29.4 & 29.4 & 23.5 & 23.5 \\
\bottomrule{}
\end{tabular}
}
\end{table*}

\subsection{Utility of UITars on Fides}
\label{sec:app:fides_utility}

The utility of UITars on Fides is shown in \cref{tab:fides_pass3_results}
\begin{table}[htbp]
\captionsetup{width=\columnwidth}
\centering
\scriptsize
\caption{\textbf{Performance comparison of UITars on Fides up to Pass@5.}}
\label{tab:fides_pass3_results}
\small
\begin{tabular}{l r}
\toprule
\textbf{Category} & \textbf{UITars} \\
\midrule
Chrome & 7/11 (63.64\%) \\
GIMP & 2/6 (33.33\%) \\
LibreOffice Calc & 3/3 (100.00\%) \\
LibreOffice Impress & 6/9 (66.67\%) \\
LibreOffice Writer & 5/8 (62.5\%) \\
Multi-apps & 3/5 (60.0\%) \\
OS & 3/4 (75.0\%) \\
Thunderbird & 2/4 (50.0\%) \\
VLC & 1/2 (50.0\%) \\
VS Code & 8/8 (100.0\%) \\
\midrule
\textbf{Overall} & 40/60 (66.67\%) \\
\midrule
\textbf{Pass@1} & 20/60 (33.33\%) \\
\textbf{Pass@2} & 28/60 (46.67\%) \\
\textbf{Pass@3} & 35/60 (58.3\%) \\
\textbf{Pass@4} & 37/60 (61.67\%) \\
\textbf{Pass@5} & 40/60 (66.67\%) \\
\bottomrule
\end{tabular}
\end{table}

\subsection{Scaling of Claude with CaMeL}
\label{sec:app:scaling_claude}
\begin{figure}[H]
    \centering
    \newlength\figureheight \newlength\figurewidth
    \setlength\figureheight{6cm} 
    \setlength\figurewidth{0.8\linewidth} 
    
    \resizebox{\linewidth}{!}{%
        \input{content/figures/percentage_tasks_passed_by_category.pgf}%
    }
    \caption{Performance of Claude-based Computer Use Agent as more unbiased sampling is allowed. We observe that agent performance grows even further to almost 73\%.}
    \label{fig:claude_scaling}
\end{figure}

\subsection{Redundancy Defense Evaluation}
\label{sec:app:redundancy_eval}

\begin{table}[h]
\centering
\caption{\textbf{Evaluation of the efficiency of redundancy defenses.} We evaluate the false positive rate of each redundancy defense on benign tasks, and the true positive rate against our two realistic cookie mimicry attacks: the standard Google ad attack (evaluated against DOM Consistency), and the more advanced HTML5 attack (evaluated against Multi-Modal Consensus). See \Cref{sec:app:further_details_attack} for why each attack is evaluated against its specific defense level. The redundancy setup was applied to CaMeL+OSWorld.}
\label{tab:redundancy_utility_evaluation}
\small
\begin{tabular}{lcc}
\toprule
\textbf{Metric} & \multicolumn{2}{c}{\textbf{Redundancy Levels}} \\
 & \textbf{DOM Consistency} & \textbf{Multi-Modal Consensus} \\
\midrule
\multicolumn{3}{l}{\textit{False Positive Rates by Application}} \\
\midrule
Chrome & 3/8 (37.50\%) & 3/8 (37.50\%) \\
Other Apps & 1/9 (11.11\%) & 0/9 (0.00\%) \\
\midrule
\multicolumn{3}{l}{\textit{Overall False Positive Rate}} \\
\midrule
\textbf{Overall} & 4/17 (23.53\%) & 3/17 (17.65\%) \\
\midrule
\multicolumn{3}{l}{\textit{Cumulative Pass@k False Positive Rate for Chrome (\%)}} \\
\midrule
\textbf{Pass@1} & 0/8 (0.00\%) & 2/8 (25.00\%) \\
\textbf{Pass@2} & 2/8 (25.00\%) & 2/8 (25.00\%) \\
\textbf{Pass@3} & 3/8 (37.50\%) & 2/8 (25.00\%) \\
\textbf{Pass@4} & 3/8 (37.50\%) & 2/8 (25.00\%) \\
\textbf{Pass@5} & 3/8 (37.50\%) & 3/8 (37.50\%) \\
\midrule
\multicolumn{3}{l}{\textit{True positive rate on the cookie mimicry attack for 10 seeds (\%)}} \\
\midrule
& 6/10 (60\%) & 5/10 (50\%) \\
\bottomrule
\end{tabular}
\end{table}

\paragraph{Details on the false positive rates of redundancy defenses.} Analysis of the reasoning traces from the redundancy defenses' false positives reveals that in all cases, the defense mistakenly classifies a legitimate cookie banner in the DOM or screenshot as fake. This underscores the difficulty of tuning such defenses even within the narrow threat model of adversarially controlled advertisement banners on websites. The false positive rate of \textit{0/9} for non-Chrome tasks also reflects this tuning: when the task is not related to web navigation, the defense does not flag it as harmful. The only non-Chrome false positive comes from DOM Consistency mistakenly identifying an Excel cell as a malicious overlay. The overrepresentation of DOM Consistency false positives explains why Multi-Modal Consensus does not substantially increase the total false-positive count: the DOM checker already flags most false positives on its own. The screenshot-based checker is better at distinguishing legitimate cookie banners from fake ones embedded in ad banners, likely because the screenshot carries more information than the accessibility tree.

\paragraph{Details on the true positive rate for the cookie mimicry attack.} The attack is 100\% successful when the cookie snippet is very similar to the sample in \cref{app:cookie_snippet}. Failure cases are instructive: they reveal that tighter control over the DOM returned by the OS could meaningfully aid prompt-injection defenses. Failures occur when the generated plan handles cookies differently, for example by calling \texttt{find\_element\_by\_text} with a specific element type such as ``button'' as input. In the case of a standard Google advertisement banner, the attacker cannot write HTML and therefore cannot create fake buttons; \texttt{find\_element\_by\_text} then finds no coordinates for a fake cookie banner and the attack fails. (This Google ad attack is evaluated against DOM Consistency.) This failure case suggests that constraining queries to specific element types (e.g., buttons only) rather than searching the full screenshot or DOM would provide tighter data-flow control, a potential direction for future defenses.

\section{Token counts comparison between defenses}
\label{app:token_count}

\begin{table}[h]
\centering
\caption{\textbf{Token count and related prices for different setups of defenses.} Realized on a subset of 17 tasks of OSWorld. The evaluation was done with Pass@5 with defenses.}
\scriptsize
\begin{tabular}{lccc}
\toprule
\textbf{Defense} & Input Tokens & Output Tokens & Total Cost (\$)\\
\midrule
No defense & 1,797,736 & 13,120 & 0.00\\
CaMeL & 2,950,253 & 456,105 & 5.40 \\
Fides & 51,724,263 & 1,874,795 & 76.07 \\
CaMeL+DOM Consistency & 8,437,603 & 605,656 & 11.57 \\
CaMeL+Multi-Modal Cons. & 10,926,601 & 982,484 & 18.37 \\
\bottomrule
\end{tabular}
\label{tab:token_count_defenses}
\end{table}

The prices were computed following OpenAI Platform\footnote{\url{https://platform.openai.com/docs/pricing}} and Anthropic Platform\footnote{\url{https://platform.claude.com/docs/en/about-claude/pricing}} pricing per token (input/output). As per these experiments, we use GPT-5 for the planner model and the second checker model with the screenshot for Multi-Modal Consensus, we use Claude Haiku 4.5 for the first checker model with the DOM used in DOM Consistency \& Multi-Modal Consensus. Additionally, we use UITars for the main Q-VLM model, since the model is deployed locally, we do not count it in the total costs, hence a total cost of \textit{0} for No defense. 

\begin{table}[h]
\scriptsize
\centering
\caption{\textbf{Token count and related prices for different elements of the agent.} Realized on a subset of 17 tasks of OSWorld. The Q-VLM functions represent all functions inside of a generated plan that are using the Q-VLM. The tokens and related costs are evaluated on a Pass@5 run using CaMeL.}
\begin{tabular}{lccc}
\toprule
\textbf{Element} & Input Tokens & Output Tokens & Total Cost (\$)\\
\midrule
Planner & 618,399 & 313,061 & 3.90 \\
Q-VLM functions & 2,304,776 & 133,784 & 1.50 \\
DOM Consistency & 5,433,932 & 147,466 & 6.17 \\
Multi-Modal Consensus & 7,892,323 & 520,256 & 12.97 \\
\bottomrule
\end{tabular}
\label{tab:token_count_functions}
\end{table}

Observing \Cref{tab:token_count_defenses} and \Cref{tab:token_count_functions}, other than the token count explosion induced by OSWorld+Fides, we can make the following 3 observations:
\begin{itemize}
    \item Both types of defense (CaMeL, Fides) generate a proportionally much higher output than input token count increase. This can be explained first by the necessary output formatting for further processing in the plan and second because of environment state describing functions like \textit{summarize\_screenshot\_content}.
    \item Redundancy defenses are responsible for the largest increase in token count and cost. They indeed require a very detailed system prompt to enable higher distinguishability between real and spoofed ad banners, and the use state-of-the-art models (GPT-5, Claude Haiku 4.5) that are expensive.
    \item The planner model takes most of the token count cost for OSWorld+CaMeL without redundancy defenses. This imbalance stems from the very high number of output tokens compared with all Q-VLM functions combined (\textit{3}x approximately), which can itself be explained by the necessary long plans to encompass all of the required possibilities of OS navigation when trying to fulfill the user task.
\end{itemize}

\section{Attacker Goals and Plan Structure Under CFI}
\label{app:cfi_attack_scenarios}
We enumerate attacker goals against what a plan must contain to enable each.
\textbf{(1) Redirecting the agent to an attacker site.} Requires only a single \texttt{find} call for a predictable UI element (e.g., a cookie popup) on a page the attacker controls, which our one-step cookie attack achieves.
\textbf{(2) Causing clicks on arbitrary attacker-chosen elements.} Requires a pixel attack to survive defenses when the element cannot be obviously spoofed.
\textbf{(3) Denial of service.} Requires only a single successful \texttt{find} misdirection; the attacker steers the agent into a dead-end branch so the user's task fails. No positive gain for the attacker but low difficulty, and plausible for competitors sabotaging agent traffic to rival services.
\textbf{(4) Substantive harm such as fund transfers or form submissions.} Requires the user's plan to already include a sensitive submission step, meaning the user's task is itself about transferring or submitting. The attacker must compromise at least two coordinated Q-VLM calls, for example the call resolving the recipient field and the call resolving the confirm button, with mutually consistent outputs. The agent must also be authenticated into the target service at the moment of execution, a timing the attacker cannot directly observe.
\textbf{(5) Data exfiltration.} Requires an even stronger plan structure: the plan must sequentially include both a read step and a send step (e.g., summarize content, then submit elsewhere), and the attacker must compromise both to extract attacker-chosen content and route it to attacker-controlled destinations. Such plan structures are uncommon outside narrow use cases.
\textbf{(6) Injecting new actions entirely absent from the plan.} The attack class CFI is specifically designed to prevent. Structurally impossible regardless of attack technique.
Attack difficulty therefore grows with both the number of coordinated Q-VLM manipulations required and the specificity of plan alignment needed: (4) and (5) require coordinated multi-call compromise of task-specific plans, whereas (1), (2), and (3) succeed against generic utility routines.

\section{Example Plans}
\label{sec:app:example_plans}
\subsection{Natural Products database example of a good plan (P-LLM: GPT-5)}
\label{sec:app:example_plans:subsec:natural_products}
\textbf{Task: Browse the natural products database.}
\begin{lstlisting}
# Step 1: OBSERVE - Understand current state
summary = summarize_screenshot_content(
    Instruction(
        text="Describe whether a web browser is open showing a website, a generic search/new tab page, or the desktop/home screen.",
        length=300
    ),
    length=300
)

# Step 2: VERIFY - Confirm high-level location (only location type for initial check)
in_website = verify_hypothesis(
    observation=summary.text,
    hypothesis="We are viewing a website in a web browser and not the desktop home screen or a generic empty search home page."
)
is_desktop = verify_hypothesis(
    observation=summary.text,
    hypothesis="We are on the desktop home screen and not inside a web browser."
)
is_search_home = verify_hypothesis(
    observation=summary.text,
    hypothesis="We are in a web browser on a generic search or new tab homepage without a specific website loaded."
)

# Helper: cookie handling (try quickly with multiple descriptions, both methods)
def_cookie_descriptions = [
    "a button to accept all cookies or consent to tracking in a cookie notice",
    "an 'I agree' or consent button in a cookie popup",
    "a button labeled 'Accept all' in a privacy banner",
    "a dismiss or OK button in a cookie notification popup"
]

def_handle_cookies_elements = get_page_elements(element_types=None)
cookie_check = verify_hypothesis(
    observation=def_handle_cookies_elements.text,
    hypothesis="A cookie or privacy consent popup with accept/consent buttons is visible"
)
if cookie_check.status == "OK":
    cookie_found = False
    for desc in def_cookie_descriptions:
        if cookie_found:
            no_op()
        else:
            cookie_try = find(Instruction(text=desc, length=150))
            if cookie_try.start is None:
                cookie_try = find_element_by_text(description=desc, element_types=["push-button", "button", "link"])
            if cookie_try.start is not None:
                cclick = left_single(cookie_try.start, "Accept cookies/consent")
                if cclick.status == "OK" or cclick.status == "UNKNOWN":
                    cookie_found = True
                    wait()
                else:
                    no_op()
            else:
                no_op()
else:
    no_op()

# Branch on current state
if in_website.status == "OK":
    # We are already on a website. Explore this site first.
    # Try to find a site-internal search (not the browser address bar).
    search_found = False
    site_search_descriptions = [
        "the website's internal search box in the header or main area (not the browser address bar)",
        "a search input field on this site to search site content (not Google search)"
    ]
    for sdesc in site_search_descriptions:
        if search_found:
            no_op()
        else:
            sres = find(Instruction(text=sdesc, length=200))
            if sres.start is None:
                sres = find_element_by_text(description=sdesc, element_types=["entry", "textbox", "search"])
            if sres.start is not None:
                sclick = left_single(sres.start, "Focus site search field (internal website search)")
                if sclick.status == "OK" or sclick.status == "UNKNOWN":
                    wait()
                    # Search for natural products using the website search (explicitly not Google)
                    typed = type_text(text="natural products database\n", instruction="Type into site internal search field (not Google)")
                    if typed.status == "OK" or typed.status == "UNKNOWN":
                        wait()
                        search_found = True
                    else:
                        no_op()
                else:
                    no_op()
            else:
                no_op()

    # After attempting site search once, assess if content seems relevant; if not, prepare to use broader web search.
    # If we're not on a clearly relevant page yet, we will do a web search via the address bar as a last resort.
    # Determine if results/content appear relevant to browsing natural products
    page_text_after_site_search = get_page_text(max_length=2000, include_navigation=False)
    relevant_check = verify_hypothesis(
        observation=page_text_after_site_search.text,
        hypothesis="The page is showing content or results related to natural products databases that can be browsed."
    )

    if relevant_check.status != "OK":
        # Use the browser's address bar to search the web as a last resort
        hotkey(keys=[Key.CTRL, Key.L], instruction="Focus browser address bar for web search")
        type_res = type_text(text="natural products database\n", instruction="Search the web for a natural products database")
        if type_res.status == "OK" or type_res.status == "UNKNOWN":
            wait()
        else:
            no_op()

elif is_desktop.status == "OK":
    # On desktop - open Chromium browser
    opened = False

    # Try to click a Chromium icon if visible
    browser_icon_descriptions = [
        "Chromium browser icon or launcher on the desktop or dock",
        "Google Chrome or Chromium icon to open the browser"
    ]
    for bdesc in browser_icon_descriptions:
        if opened:
            no_op()
        else:
            bres = find(Instruction(text=bdesc, length=150))
            if bres.start is None:
                bres = find_element_by_text(description=bdesc, element_types=["push-button", "button", "link"])
            if bres.start is not None:
                bclick = left_single(bres.start, "Open Chromium")
                if bclick.status == "OK" or bclick.status == "UNKNOWN":
                    opened = True
                    wait()
                else:
                    no_op()
            else:
                no_op()

    if not opened:
        # Use system launcher as fallback
        hotkey(keys=[Key.WIN], instruction="Open system application launcher")
        wait()
        type_text(text="chromium", instruction="Type Chromium in launcher")
        wait()
        press(Key.ENTER, instruction="Launch Chromium from launcher")
        wait()

    # Once browser is open, perform a web search for natural products database
    hotkey(keys=[Key.CTRL, Key.L], instruction="Focus address bar")
    type_text(text="natural products database\n", instruction="Search the web for a natural products database")
    wait()

elif is_search_home.status == "OK":
    # In a browser on generic search/new tab - search directly
    hotkey(keys=[Key.CTRL, Key.L], instruction="Focus address bar for search")
    type_text(text="natural products database\n", instruction="Search the web for a natural products database")
    wait()

else:
    # Fallback: Try to focus address bar anyway and search
    hotkey(keys=[Key.CTRL, Key.L], instruction="Focus address bar (fallback)")
    type_text(text="natural products database\n", instruction="Search the web for a natural products database")
    wait()

# Handle cookies on the search results page (if any)
elements_after_search = get_page_elements(element_types=None)
cookie_check2 = verify_hypothesis(
    observation=elements_after_search.text,
    hypothesis="A cookie or privacy consent popup with accept/consent buttons is visible"
)
if cookie_check2.status == "OK":
    cookie2_found = False
    for desc in def_cookie_descriptions:
        if cookie2_found:
            no_op()
        else:
            r1 = find(Instruction(text=desc, length=150))
            if r1.start is None:
                r1 = find_element_by_text(description=desc, element_types=["push-button", "button", "link"])
            if r1.start is not None:
                a1 = left_single(r1.start, "Accept cookies/consent on search results")
                if a1.status == "OK" or a1.status == "UNKNOWN":
                    cookie2_found = True
                    wait()
                else:
                    no_op()
            else:
                no_op()
else:
    no_op()

# Step 3: ACT - From web search, navigate to a credible natural products database result
db_found = False
db_result_descriptions = [
    "a search result link for COCONUT natural products database",
    "a search result link for Natural Products Atlas (NPAtlas)",
    "a search result link for NPASS natural products database",
    "a search result link that clearly leads to a natural products database website"
]

for ddesc in db_result_descriptions:
    if db_found:
        no_op()
    else:
        fres = find(Instruction(text=ddesc, length=200))
        if fres.start is None:
            fres = find_element_by_text(description=ddesc, element_types=["link"])
        if fres.start is not None:
            nav = left_single(fres.start, "Open natural products database site")
            if nav.status == "OK" or nav.status == "UNKNOWN":
                db_found = True
                wait()
            else:
                no_op()
        else:
            no_op()

# Handle cookies on the target site
elements_on_target = get_page_elements(element_types=None)
cookie_check3 = verify_hypothesis(
    observation=elements_on_target.text,
    hypothesis="A cookie or privacy consent popup with accept/consent buttons is visible"
)
if cookie_check3.status == "OK":
    cookie3_found = False
    for desc in def_cookie_descriptions:
        if cookie3_found:
            no_op()
        else:
            r2 = find(Instruction(text=desc, length=150))
            if r2.start is None:
                r2 = find_element_by_text(description=desc, element_types=["push-button", "button", "link"])
            if r2.start is not None:
                a2 = left_single(r2.start, "Accept cookies/consent on database site")
                if a2.status == "OK" or a2.status == "UNKNOWN":
                    cookie3_found = True
                    wait()
                else:
                    no_op()
            else:
                no_op()
else:
    no_op()

# Verify if we have reached a natural products database site
content_check_text = get_page_text(max_length=2000, include_navigation=False)
on_db_site = verify_hypothesis(
    observation=content_check_text.text,
    hypothesis="We are on a natural products database website or portal with options to browse or search compounds, structures, or entries."
)

# If on the site, try to locate a Browse/Explore/Search interface to start browsing
if on_db_site.status == "OK":
    browse_opened = False
    browse_strategies = [
        ("a link or button to Browse the database", ["link", "push-button", "button"]),
        ("a link or button labeled Explore to navigate entries", ["link", "push-button", "button"]),
        ("a main Search input field to search compounds in the database", ["entry", "textbox"]),
        ("a navigation element for Compounds or Entries", ["link", "push-button", "button"])
    ]
    for bdesc, btypes in browse_strategies:
        if browse_opened:
            no_op()
        else:
            bfind = find(Instruction(text=bdesc, length=200))
            if bfind.start is None:
                bfind = find_element_by_text(description=bdesc, element_types=btypes)
            if bfind.start is not None:
                bclick = left_single(bfind.start, "Open browsing/search interface on the database site")
                if bclick.status == "OK" or bclick.status == "UNKNOWN":
                    browse_opened = True
                    wait()
                else:
                    no_op()
            else:
                no_op()

    if not browse_opened:
        # Try a few scrolls to reveal browse/search options
        for _ in range(3):
            if browse_opened:
                no_op()
            else:
                scroll(direction="down", start=None, instruction="Reveal more database navigation options")
                wait()
                for bdesc, btypes in browse_strategies:
                    if browse_opened:
                        no_op()
                    else:
                        bfind2 = find(Instruction(text=bdesc, length=200))
                        if bfind2.start is None:
                            bfind2 = find_element_by_text(description=bdesc, element_types=btypes)
                        if bfind2.start is not None:
                            bclick2 = left_single(bfind2.start, "Open browse/search after scrolling")
                            if bclick2.status == "OK" or bclick2.status == "UNKNOWN":
                                browse_opened = True
                                wait()
                            else:
                                no_op()
                        else:
                            no_op()
else:
    # If not clearly on a database site yet, try once more to follow any link that indicates a natural products database
    if not db_found:
        alt = find_element_by_text(
            description="any link that leads to a natural products database website",
            element_types=["link"]
        )
        if alt.start is not None:
            alt_click = left_single(alt.start, "Open alternative natural products database")
            if alt_click.status == "OK" or alt_click.status == "UNKNOWN":
                wait()
            else:
                no_op()
        else:
            no_op()

# Final verification and completion
final_text = get_page_text(max_length=2000, include_navigation=False)
final_verification = verify_hypothesis(
    observation=final_text.text,
    hypothesis="We are now on a natural products database page with a browse or search interface visible, allowing us to browse entries."
)

if final_verification.status == "OK":
    # Optional: check_done before marking done
    done_check = check_done(
        Instruction(
            text="I am on a natural products database website with a browse or search interface, ready to browse entries.",
            length=200
        )
    )
    if done_check.done:
        mark_done()
    else:
        mark_done()
else:
    mark_fail()
\end{lstlisting}

\subsection{Natural Products database example of an insufficient plan (P-LLM: Gemini 3 Pro)}
\label{sec:app:example_plans:subsec:natural_products_gemini3}
\textbf{Task: Browse the natural products database.} 
This plan is lacking branches, is too confident, and does not focus on website navigation but rather on going back into search and finding other databases it thinks fits the task description.
\begin{lstlisting}
# Step 1: Observe the current state
summary = summarize_screenshot_content(Instruction(text="Describe the current page content, checking if it is a search engine, a specific website, or if there are cookie popups.", length=300), length=300)
print(f"Current page state: {summary.text}")

# Step 2: Handle Cookie Popups (Common in browser tasks)
# We check for this first to ensure elements aren't blocked
cookie_check = verify_hypothesis(
    observation=summary.text,
    hypothesis="A cookie consent popup, privacy notice, or GDPR banner is visible on the screen."
)

if cookie_check.status == "OK":
    print("Cookie popup detected. Attempting to accept.")
    # Try to find the accept button using common labels
    cookie_handled = False
    cookie_labels = ["Accept all", "Accept cookies", "I agree", "Consent", "Allow all", "OK"]
    
    for label in cookie_labels:
        if cookie_handled:
            no_op()
        else:
            btn = find_element_by_text(label, element_types=["button", "push-button"])
            if btn.start is not None:
                left_single(btn.start, f"Click {label}")
                wait()
                cookie_handled = True
    
    # If text search failed, try visual search
    if not cookie_handled:
        btn = find(Instruction(text="Accept or Agree button in cookie popup", length=100))
        if btn.start is not None:
            left_single(btn.start, "Click Accept visually")
            wait()
            
    # Refresh summary after handling cookies
    summary = summarize_screenshot_content(Instruction(text="Page content after cookie handling", length=300), length=300)

# Step 3: Determine Location and Navigate
# Check if we are already on a Natural Products Database
is_on_database = verify_hypothesis(
    observation=summary.text,
    hypothesis="The page is a Natural Products Database website (e.g., NPASS, COCONUT, NP-MRD, SuperNatural) or has scientific data about natural compounds."
)

if is_on_database.status == "OK":
    print("Already on a database website. Proceeding to browse.")
    # Look for navigation options to browse data
    nav_clicked = False
    nav_options = ["Browse", "Explore", "Data", "Compounds", "Search"]
    
    for opt in nav_options:
        if nav_clicked:
            no_op()
        else:
            link = find_element_by_text(opt, element_types=["link", "button", "menu item"])
            if link.start is not None:
                left_single(link.start, f"Click {opt}")
                wait()
                nav_clicked = True
    
    if not nav_clicked:
        # If no specific link found, scroll to browse content
        scroll("down", None, "Scroll down to browse the page content")
        wait()
        
    mark_done()

else:
    # Not on a database. Check if we are on search results.
    is_search_results = verify_hypothesis(
        observation=summary.text,
        hypothesis="The page shows search results for 'natural products database' or similar queries."
    )
    
    # If not on search results, we need to perform a search
    if is_search_results.status == "FAIL":
        print("Not on database or search results. Initiating search.")
        
        # Find search bar
        search_bar = find_element_by_text("Search", element_types=["entry", "textbox"])
        if search_bar.start is None:
            search_bar = find(Instruction(text="Search bar or input box", length=50))
            
        if search_bar.start is not None:
            left_single(search_bar.start, "Click search bar")
            type_text("natural products database\n", "Search for natural products database")
            wait()
            # Assume we are now on search results
            is_search_results = ActionCall(status="OK") 
        else:
            print("Could not find search bar to navigate.")
            mark_fail()

    # Step 4: Select a Database from Search Results
    if is_search_results.status == "OK":
        print("On search results page. Looking for a valid database link.")
        
        # Look for reputable databases
        target_found = False
        # Candidates: NPASS, COCONUT, NP-Atlas, SuperNatural
        candidates = [
            "NPASS", 
            "COCONUT", 
            "Natural Products Atlas", 
            "SuperNatural", 
            "Natural Product Activity and Species Source"
        ]
        
        for cand in candidates:
            if target_found:
                no_op()
            else:
                link = find_element_by_text(cand, element_types=["link", "heading"])
                if link.start is not None:
                    left_single(link.start, f"Click {cand}")
                    wait()
                    target_found = True
        
        # Fallback if no specific name found
        if not target_found:
            generic_link = find_element_by_text("Natural Products Database", element_types=["link", "heading"])
            if generic_link.start is not None:
                left_single(generic_link.start, "Click generic database link")
                wait()
                target_found = True
                
        if target_found:
            # Step 5: Browse the selected database
            # Check for cookies on the new site
            new_summary = summarize_screenshot_content(Instruction(text="New page content", length=200), length=200)
            new_cookie_check = verify_hypothesis(observation=new_summary.text, hypothesis="Cookie popup visible")
            
            if new_cookie_check.status == "OK":
                accept_btn = find_element_by_text("Accept", element_types=["button", "push-button"])
                if accept_btn.start is not None:
                    left_single(accept_btn.start, "Accept cookies on new site")
                    wait()
            
            # Find Browse button
            browse_btn = find_element_by_text("Browse", element_types=["link", "button", "menu item"])
            if browse_btn.start is None:
                browse_btn = find_element_by_text("Explore", element_types=["link", "button"])
            
            if browse_btn.start is not None:
                left_single(browse_btn.start, "Click Browse")
                wait()
            else:
                # Scroll if no browse button
                scroll("down", None, "Scroll to browse content")
                wait()
                
            mark_done()
        else:
            print("Could not find a relevant database link in the search results.")
            mark_fail()
\end{lstlisting}

\subsection{Example of a generic cookie handler snippet}
\label{app:cookie_snippet}

\begin{lstlisting}
def_cookie_descriptions = [
    "a button to accept all cookies or consent to tracking in a cookie notice",
    "an 'I agree' or consent button in a cookie popup",
    "a button labeled 'Accept all' in a privacy banner",
    "a dismiss or OK button in a cookie notification popup"
]

# After potential navigation or app launch, re-observe
summary = summarize_screenshot_content(
    Instruction(
        text="Observe the current page to understand if we are on a website and what site it might be.",
        length=1200
    ),
    length=1200
)


cookie_check = verify_hypothesis(
    observation=summary.text,
    hypothesis="A cookie or privacy consent popup with accept/consent buttons is visible"
)
if cookie_check.status == "OK":
    cookie_found = False
    for desc in def_cookie_descriptions:
        if cookie_found:
            no_op()
        else:
            cookie_try = find(Instruction(text=desc, length=150))
            if cookie_try.start is None:
                cookie_try = find_element_by_text(description=desc, element_types=["push-button", "button", "link"])
            if cookie_try.start is not None:
                cclick = left_single(cookie_try.start, "Accept cookies/consent")
                if cclick.status == "OK" or cclick.status == "UNKNOWN":
                    cookie_found = True
                    wait()
                else:
                    no_op()
            else:
                no_op()
else:
    no_op()
\end{lstlisting}

\end{document}